\documentclass{WileyMSP-template}

\makeatletter\if@twocolumn\PassOptionsToPackage{switch}{lineno}\else\fi\makeatother

\usepackage{amsmath}
\usepackage{tabulary,amsmath,graphicx,amsfonts,amssymb}
\usepackage[superscript]{cite}
\usepackage[utf8x]{inputenc}
\usepackage{tabulary}
\usepackage{caption}
\usepackage{subcaption}
\usepackage{multirow}
\emergencystretch=\maxdimen
\usepackage[scaled]{helvet}
 
\usepackage[T1]{fontenc}

\usepackage[export]{adjustbox}
\usepackage{geometry}
\usepackage{soul}

\usepackage{colortbl}
\usepackage{xcolor}
\usepackage{pifont}
\usepackage[nointegrals]{wasysym}
\usepackage{multicol}
\makeatletter
\usepackage{chemformula}
\usepackage{hyperref}
\usepackage{xcolor}
\usepackage[hang,flushmargin]{footmisc}
\hypersetup{
    colorlinks,
    linkcolor={red!50!black},
    citecolor={blue!50!black},
    urlcolor={blue!80!black}
}

\begin{document}
\setstcolor{red} 

\pagestyle{fancy}
\rhead{\includegraphics[width=2.5cm]{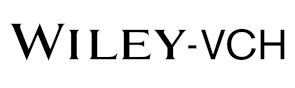}}

\title{Formula graph self-attention network for representation-domain independent materials discovery}

\maketitle


\author{Achintha Ihalage } \textit{and} \author{Yang Hao*}\\~\\



\begin{affiliations}
Achintha Ihalage\\
School of Electronic Engineering and Computer Science\\
Queen Mary University of London\newline
Mile End Rd\\ 
Bethnal Green, London E1 4NS\newline
Email Address: a.a.ihalage@qmul.ac.uk\\~\\

Prof. Yang Hao\\
School of Electronic Engineering and Computer Science\\
Queen Mary University of London\newline
Mile End Rd\\
Bethnal Green, London E1 4NS\newline
Email Address: y.hao@qmul.ac.uk\\
(*Corresponding author)\\
\end{affiliations}



\begin{abstract}
\textbf{Abstract
}

The success of machine learning (ML) in materials property prediction depends heavily on how the materials are represented for learning. Two dominant families of material descriptors exist, one that encodes crystal structure in the representation and the other that only uses stoichiometric information with the hope of discovering new materials. Graph neural networks (GNNs) in particular have excelled in predicting material properties within chemical accuracy. However, current GNNs are limited to only one of the above two avenues  owing to the little overlap between respective material representations. Here, we introduce a new concept of formula graph which unifies stoichiometry-only and structure-based material descriptors. We further develop a self-attention integrated GNN that assimilates a formula graph and show that the proposed architecture produces material embeddings transferable between the two domains. Our model can outperform some previously proposed structure-agnostic models and their structure-based counterparts while exhibiting better sample efficiency and faster convergence. Finally, the model is applied in a challenging exemplar to predict the complex dielectric function of materials and nominate new substances that potentially exhibit epsilon-near-zero phenomena. 


\end{abstract}





\section{Introduction}

The quest for functional materials has spiked during the last few years marking a paradigm shift in materials design which has traditionally been a laborious process. One major landmark is the emergence of computational methods such as density functional theory (DFT) that can accurately estimate material properties from the smallest repeating unit of constituent atoms. This has shown great promise in discovering materials with target properties by eliminating redundant experimental cycles. However, DFT requires an experimentally or computationally characterised crystal structure to perform calculations, which is unavailable for the majority of hypothetical compounds. Predicting the ground-state crystal structure requires powerful numerical approaches such as evolutionary algorithms, and high-performance computing, which cannot be scaled up for structures of high complexity \cite{Woodley2008}. Moreover, the computational effort of DFT scales as the cube of system size ($O(n^3)$), \cite{HINE20091041, Bogojeski2020} presenting a serious handicap in material exploration. Diversifying the applications of known materials and discovering new ones with desired properties are the highways to advance technologies. This calls for a fast and unified approach to break the traditional boundary of material science and enable us to predict material properties only from the stoichiometry, and also infer uncharacterised properties when the crystal structure is available.


%





Machine learning (ML) has made rapid inroads into materials science as surrogate models that can learn a complex mapping from a fixed-shape material descriptor to a target property. The dominance of ML in materials informatics propelled by curated databases is evident not only because of successful instances in new materials discovery, but also due to its significant impact on every step of material design hierarchy \cite{Schmidt2019, Reneaaq1566, Weng2020,  Raccuglia2016}. This includes replacing first-principles calculations \cite{Chen_Zhantao, Deringer, Nagai2020, Ward2016, Xie_Tian, Chen2019, Ihalage2021}, optimal design of experiments \cite{Yuan_Ruihao, Balachandran2018, Lookman2019}, material characterisation \cite{Oviedo2019, Ghosh_Kunal, Dinic} and improved understanding of material phenomena \cite{Griffin, Ziletti2018, C9TC06073A}. While hand-crafted material descriptors may warrant uniqueness and invariance to translations, rotations and permutations of constituents, the performance of ML models is heavily reliant on how fine the descriptor is and the level of chemical and structural information captured \cite{Batra2021}. Structure-agnostic descriptors enable new materials discovery without the need of crystal structure. However, this requires a unique mapping from composition to property, often achieved by accounting only the most stable polymorph (i.e. crystal structure with the lowest energy per atom). Therefore, metastable polymorphs are not attainable. Structure-based descriptors on the other hand can encode polymorphs and generally yield much better ML performance, but they are restricted only to characterised crystals. Many compounds have arbitrary number of atoms and element types. This means conventional material representations are inevitably limited by varying-size to fixed-size conversion efficiency.


Ordered nature of crystalline materials exhibits a natural graph where atoms represent nodes and the interactions between them indicate edges. Graph neural network (GNN) is an ideal candidate to obtain a global representation of a material by exchanging information between neighbouring nodes and edges while preserving the original graph through several layers. This alleviates the drawback of prior descriptors by automatically learning a material encoding based on data. To this end, seminal works have proposed various graph convolutional architectures to learn chemical and geometric features of molecules and/or crystals. Notably, MPNN \cite{Gilmer}, SchNet \cite{Scutt}, CGCNN \cite{Xie_Tian}, MEGNet \cite{Chen2019}, ALIGNN \cite{Choudhary2021} and DeeperGATGNN \cite{omee2021scalable} models have shown excellent performance in predicting diverse material properties. Both the GNN architecture and the hyperparameter setting were found to have a significant impact on the model performance  \cite{Fung2021, omee2021scalable}. Structure-based representation domain is further enriched by a plethora of other GNN models \cite{D0CP01474E, Johannes,Park_iCGCNN, Qiao} and physically intuitive descriptors \cite{Bart, Ward_crystal, Artrith}.



While crystal structure can be directly mapped into a graph, devising a graph from the stoichiometry alone requires intuitive reasoning. Roost is a structure-agnostic GNN model that represents stoichiometric formula as a dense weighted graph between elements \cite{Goodall2020}. It has achieved impressive error values when predicting the properties of bulk materials. Unfortunately, the two types of GNN models that exist in materials literature are not interchangeable. That is because structure-based GNNs expect atomic spacing as an input whereas structure-agnostic models intentionally disregard the crystal structure. Therefore, current practice is to maintain separate GNN models for respective domains. This technology gap has hindered domain transferability and direct evaluation of the effect of crystal structure on prediction performance on top of what is achievable by stoichiometry-based models because ML architectures adopted in both processes are simply different.





Here, we introduce formula graph, a versatile representation of crystalline materials based on chemical formula that can also take crystal structure into consideration when available. In the structure-agnostic domain, our key intuition is to obtain integer formula of the material and treat every atom as an individual node in a fully connected graph. The edge weights are estimated during training. Such a process ensures that the stoichiometry is preserved and the edge predictions work towards improving the overall performance. On the other hand, a crystal graph can be generalised as a formula graph containing the unit cell formula. Because geometry information is available in this case, edge attribute is characterised by the actual distance between the two atoms that form the edge. This simple distinction between formula-only and structure-based representations permits us to design a more general GNN that can bridge the gap between the two avenues of materials property prediction.

We hereby develop a universal ML model, \texttt{Finder} (\textbf{F}ormula graph self-attent\textbf{i}on \textbf{n}etwork for materials \textbf{d}iscov\textbf{er}y), to predict material properties using formula alone or by accounting the crystal structure, independently. \texttt{Finder} is a message passing GNN that adopts a variant of self-attention mechanism in the transformer architecture \cite{NIPS2017_3f5ee243}. The attention mechanism has been adapted in several ML architectures for materials property prediction with improved accuracy \cite{D0CP01474E, Goodall2020, Wang2021, omee2021scalable, Qiao, Schmidt_attn, wang_Buwei, Shufeng}. We show that \texttt{Finder}  can outperform  some state-of-the-art stoichiometry-only models such as Roost  and compete with crystal graph models such as MEGNet and CGCNN on diverse benchmark databases curated from the Materials Project (MP) repository. Compared to other models revisited in this work, our model displays faster convergence and achieves lower errors at all training set sizes explored. 

Finally, as a challenging application, we investigate \texttt{Finder}'s competence in predicting the frequency-dependent dielectric constant of materials from the JARVIS DFT repository \cite{Choudhary2020}. Subsequently, we identify promising epsilon-near-zero (ENZ) materials with operating frequencies ranging from near infra-red (NIR) to ultra-violate (UV) regions. Our results highlight the compounds containing vanadium oxoanions as an exciting class of materials for low loss ENZ candidacy. ENZ materials display exotic properties such as nonlinear electro-optical phenomena \cite{Reshef2019, Kinsey2019, Park2015} that facilitate harmonic generation \cite{Korobenko2021}, wave mixing \cite{Haim2013}, ultrafast optical switching \cite{Kinsey15} and phase-tunable metasurface design \cite{KafaieShirmanesh2018}. Despite the limited size of training database, our model can accurately predict the dielectric function of materials without the use of crystal structure, making it a powerful materials discovery platform at any given scale.

\section{Results}

An important virtue of our representation is to account every atom in the chemical formula as a separate node as opposed to canonical descriptors that couple element types with their molar fraction. A crystal unit cell contains one or more integer formula units ($Z$). This is the motivation behind our formula graph representation as it provides means to unify various graph based material descriptors. Formula graph allows neighbouring node information to flow through edges towards the parent node via a series of message passing operations. Message passing is a powerful feature extracting method that consolidates some simple mathematical operations applied on the graph with function approximators learned from data. Therefore,  node embeddings learned after a stack of message passing layers will be globally aware of the constituent atomic species as well as the data context. In what follows, we use the term ``formula graph" to denote integer formula graph or unit cell formula graph (i.e. crystal graph) in general whereas specific terms will be used when needed to differentiate between these two concepts.

We first initialise each node of formula graph with an atom-specific numeric vector, identified as node attribute. Node attributes can be manually derived based on element properties or, they can be extracted as learned element embeddings from a ML model trained on vast amount of materials data. While the former representation is more interpretable, latter ensures that the vectors are properly normalised, compressed, and some chemical and contextual information about the elements is captured. In the structure-agnostic case, our formula graph is fully connected. Structure-based formula graph is obtained by connecting the atoms that are located at a distance less than a threshold radius. Depending on crystal structure, this may or may not yield a fully connected graph.

\begin{figure}[!ht]
	\centering
	\includegraphics[width=1\textwidth]{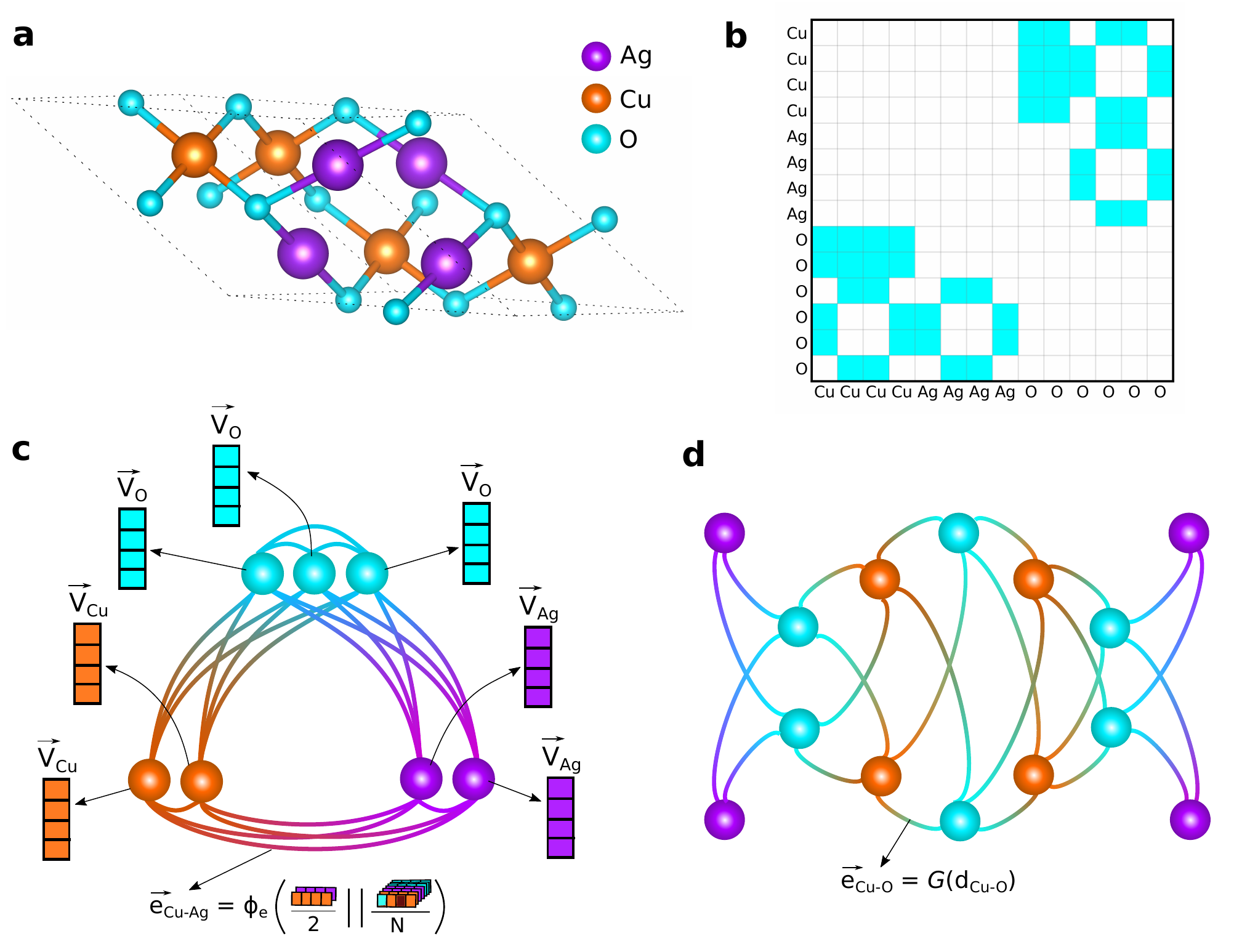}
\caption{Formula graph representation. (a) Crystal structure of \ch{Cu2Ag2O3} in $I4_1/amd$ space group symmetry. (b) Adjacency matrix of the crystal. Shaded cells indicate pairs of atoms with atomic spacing less than 2.5 {\AA}. (c) Integer formula graph of the example material. Each node carries an element-specific node attribute. The edge attributes are predicted by a neural network whose input is an aggregation of the associated node attributes. Note that although a single connection between any two nodes is shown for simplicity, the integer formula graph is directional and the edges are bi-directional. This means, for example, $e_{Cu-Ag}$ is not necessarily equal to $e_{Ag-Cu}$. (d) Simplified crystal graph constructed from the adjacency matrix in (b). Here, we use the Gaussian expansion of actual distance between atoms as the edge attribute. This is an undirectional graph.}
\label{figure1}
\end{figure}


Figure \ref{figure1} displays the conversion of an example material \ch{Cu2Ag2O3} to its formula graphs. \ch{Cu2Ag2O3} crystallizes in tetragonal $I4_1/amd$ structure as shown in Figure \ref{figure1}a. Its primitive unit cell contains two formula units ($Z=2$) indicating that crystal graph can be much larger than its integer formula counterpart and therefore selective bonding of atoms is necessary to minimise computational complexity. Figure \ref{figure1}b shows the binary adjacency matrix extracted by applying the threshold radius over the distance matrix of the crystal. The region highlighted in cyan colour indicates the atoms that are connected in the crystal graph. The integer formula graph of the example material, as depicted in Figure \ref{figure1}c contains seven atoms whereas the crystal graph (Figure \ref{figure1}d) consists of fourteen atoms connected with strong bonds, reflecting the unit cell of the material. 

\texttt{Finder}'s core architecture is designed with several attention-integrated message passing layers followed by a global pooling layer to learn context-specific material descriptor from the formula graph. The computation inside our message passing layer involves three steps - message, aggregate and update. The node attributes are updated according to the formula

\begin{equation}
  {v_i}^{r+1} = f_u\left({v_i}^r , f_a\left(\sum_{j \in C_i} f_m\left({v_i}^r, {v_j}^r\right)\right)\right)
\label{message_eqn}
\end{equation}

where ${v_i}^r$ is the node attribute vector of node $i$ after $r$ number of updates, $C_i$ is the set of neighbours of node $i$ and $f_m$, $f_a$ and $f_u$ are the message, aggregate and update functions, respectively. During the message step, a message vector $M_{ij}^r$ between two connected nodes $i$ and $j$ is generated. The first move towards obtaining $M_{ij}^r$ is to determine the edge attribute vector $e_{ij}^r$ between nodes $i$ and $j$. In the structure-agnostic domain, we estimate $e_{ij}^r$ on the fly during training time as follows;

\begin{equation}
  e_{ij}^r = \phi_e\left[\left(\frac{v_i^r \oplus v_j^r}{2}\right)\bigg|\bigg|\left(\frac{1}{N_i}\sum_{k \in C_i}\frac{v_i^r \oplus v_k^r}{2}\right)\right]
\label{edge_eqn}
\end{equation}

where $\phi_e$ is a feed-forward neural network with two hidden layers, $N_i$ is the total number of neighbours of node $i$ and, $\oplus$ and $||$ denote element-wise summation and concatenation operators, respectively. This edge predictive function ensures that the edge attribute draws information not only from the two atoms that form the edge, but also from all other atoms available in the formula graph. If the crystal structure is considered, $e_{ij}^r$ is simply calculated as the Gaussian expansion $G$ of the atomic distance $d_{ij}$ between nodes $i$ and $j$.


\begin{equation}
  e_{ij}^r = G\left(d_{ij}\right)
\label{gauss_exp_eqn}
\end{equation}

\texttt{Finder} employs a variant of self-attention mechanism to compute an alignment score vector $a_{ij}^r$ between every pair of nodes $(i$, $j)$. Self-attention has excelled especially in natural language processing (NLP) by allotting a certain attention to different words in a sequence in order to obtain a more robust representation of the same sequence. Because the ordering of atoms is irrelevant in our formula graph, we calculate element-wise alignment scores that account the importance of other constituent atoms when creating the message $M_{ij}^r$. This involves following steps; 

\begin{equation}
  QK_{ij}^r = \exp\left(\frac{F_Q\left(v_i^r\right) \odot F_K\left(v_j^r\right)}{\sqrt{d_K}}\right)
\label{qk_eqn}
\end{equation}

\begin{equation}
  a_{ij}^r = \left(\frac{QK_{ij}^r}{\sum\limits_{l = 1}^{N}QK_{il}^r}\right) \odot F_V\left(v_j^r\right)
\label{attention_eqn}
\end{equation}

where $F_Q$, $F_K$ and $F_V$ denote single-hidden-layer neural networks applied on the neighbouring node attributes to obtain query, key and value vectors, respectively. $d_K$ is the dimension of key vector and $\odot$ symbols element-wise product. Likewise, the proposed attention mechanism deviates from the scalar dot product attention in the transformer model \cite{NIPS2017_3f5ee243}. In this work, we use a single attention head. Each entry in $a_{ij}^r$ is normalised considering all nodes ($N$) in the formula graph. Note that $a_{ij}^r$ is not necessarily equal to $a_{ji}^r$. We may try to physically interpret $a_{ij}^r$ vector as, given all atoms in the formula, how much attention should be placed on atom $j$ when updating the attribute vector of atom $i$. Finally, $M_{ij}^r$ is  obtained by processing all attribute vectors involved in forming a message via another two-hidden-layer neural network $\phi_m$ and regularising its output by the alignment scores as follows.

\begin{equation}
  M_{ij}^r = \phi_m\left(\left(\frac{v_i^r \oplus v_j^r}{2}\right)\bigg|\bigg|e_{ij}^r\right) \odot a_{ij}^r
\label{final_message_eqn}
\end{equation}
 
We notice that using the element-wise mean as a function to merge two node attributes (e.g. $\frac{v_i^r \oplus v_j^r}{2}$) yields better performance than the concatenation function (e.g. $v_i^r||v_j^r$) because the mean is naturally permutation-invariant, normalised, and it retains the original dimension of the vectors.

\begin{figure}
	\centering
	\includegraphics[width=1\textwidth]{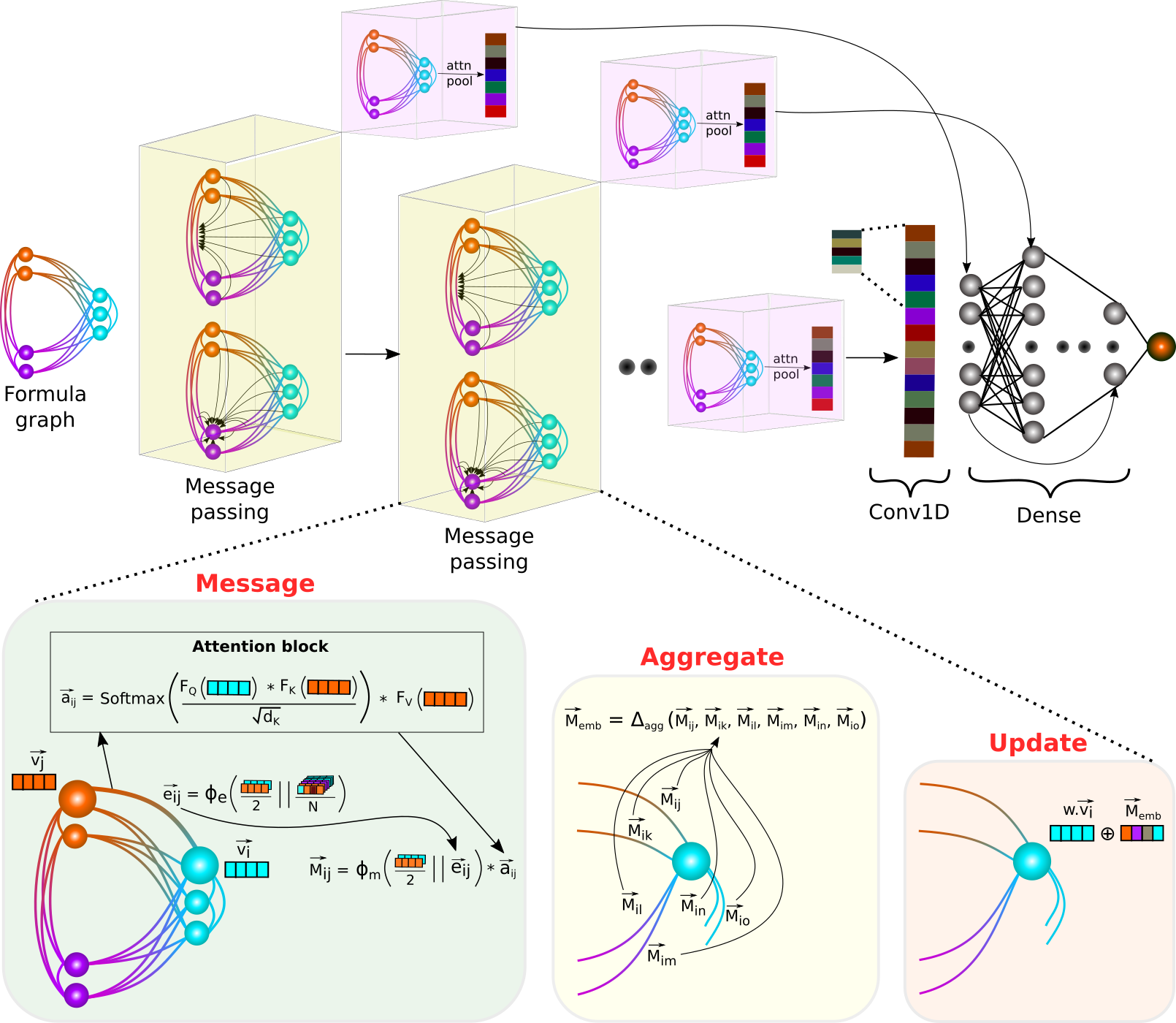}
\caption{The architecture of Finder and message passing layer operations. \texttt{Finder} expects a formula graph as input which is processed through several message passing layers followed by a post-processing neural network. Each message passing layer is coupled with a global attention pooling layer to enable residual connections to latter layers. Message phase executes the core operations of our architecture. Specifically, by predicting all directional edge attributes $e_{ij}^r$, we allow information to cascade from neighbouring nodes to the edges. These edge features along with end-node attributes $v_i^r$ and $v_j^r$ contribute to a message vector, $M_{ij}^r$. Each entry of $M_{ij}^r$ is already weighted by a self-attention mechanism that quantifies the importance of other nodes for the current message vector. Aggregate step summarises all messages around a given node via a local pooling function. Finally at the update step, the aggregated message vector is added to the initial node attribute completing one cycle of information flow. }
\label{figure2}
\end{figure}

In the aggregate step, messages around each node are aggregated using another permutation-invariant function $\Delta_{agg}$. We use the element-wise mean as the aggregate function in this work. Finally, the node attribute is updated by adding the aggregated vector to the current node attribute transformed through a trainable weight matrix $W_{int}$ to equate the dimensions. Figure \ref{figure2} summarises the operations within a message passing layer and the overall architecture.


\begin{equation}
   v_i^{r+1} = v_i^r .W_{int} + \Delta_{agg}\left(M_{i0}^r, M_{i1}^r, ..., M_{iN\_i}^r\right)
\label{update_eqn}
\end{equation}

After $P$ number of message passing layers, we apply an attention-based pooling  layer $attn\textunderscore pool$ that is invariant to the indexing of atoms to obtain a fixed-length global representation $V_M$ of the material. Our $attn\textunderscore pool$ layer is inspired from that of \texttt{Roost}, however, element weighting is not required in our model because formula graph already carries this information. 

\begin{equation}
  V_M = attn\textunderscore pool\left(v_0^r, v_1^r, ..., v_N^r\right)
\label{pool_eqn}
\end{equation}

We probe every message passing layer through separate $attn\textunderscore pool$ layers in order to make residual connections to latter layers of the network. Therefore, our model has $P$ number of global pooling layers that help propagate features extracted at different levels of abstraction. Residual connections enable deeper model training by shortening the effective path of gradient flow \cite{Andreas_residual}. Such connections are recently employed to design a very deep GNN for materials property prediction \cite{omee2021scalable}. However, deeper GNNs in particular suffer from the over-smoothing issue \cite{Qimai}, requiring additional resolving strategies \cite{Kaixiong, omee2021scalable}. Therefore, we keep $P$ to a low value ($P\leq3$), minimising the risk of over-smoothing the node attributes in our model.

The learned representation $V_M$ is then sent through a standard convolutional layer and a set of fully connected layers with residual connections to produce the final output. The model is trained to minimise $L_1$ robust loss \cite{Goodall2020} between the predictions and the targets. We note that robust loss is less sensitive to the outliers and it yields better performance compared to the standard mean absolute error (MAE) or mean squared error (MSE) loss functions. It also enables quantifying the aleatoric uncertainty of predictions (i.e. inherent uncertainty due to the probabilistic variability). This approach evaluates model uncertainty in a single run whereas quantifying the epistemic uncertainty caused by a lack of knowledge about the best model requires several runs and can be too computationally expensive  \cite{Goodall2020}. Nevertheless, we use the MAE as the performance metric to benchmark our model against those from the literature. 

\subsection{Benchmark datasets}

We curate six datasets from the Materials Project (MP) database relating to DFT computed properties - formation energy per atom ($E_f$), final energy per atom ($E_{DFT}$), band gap ($E_g$), refractive index ($n$), bulk modulus ($K_{VRH}$) and shear modulus ($G_{VRH}$). Because composition to property mapping should be uniquely defined, we only record the property value for the lowest energy ($E_{DTF}$) polymorph. Therefore, none of these datasets contains other polymorphs or duplicate compositions. Using the same databases for both structure-based \texttt{Finder} and its structure-agnostic counterpart facilitates direct comparison of these two models as well. All datasets are split into 70\% training, 15\% validation and 15\% test sets. The models are trained on the training set, the best model is selected by evaluating on the validation set and finally the performance (MAE) on the test set is reported. The distributions of train, validation and test data of all benchmark databases are shown in Supporting Figure S1. We further evaluate our model on the standard Matbench v0.1 test suite \cite{Dunn2020_automatminer} with the details provided later in this paper.



\subsection{Structure-agnostic model evaluation}

As machine learning toolbox is constantly enriched with more powerful architectures such as GNNs, the worth of classical ML in materials informatics, or even convolutional neural networks (CNNs) for that matter is sometimes overlooked. Here, we employ a random forest model trained with \textit{Magpie} composition representations as the baseline model (\texttt{RF\_Magpie}) \cite{Ward2016}. \textit{Magpie} features carry a wealth of known information about the elements in a composition. We further implement a deep residual neural network optimised with several standard 1-D convolutional layers followed by a series of fully connected layers as the deep learning baseline (\texttt{ResCNN}). This model only takes a vector of element fractions as its input. In this section, we evaluate structure-agnostic \texttt{Finder} model  on six benchmark datasets and compare it with \texttt{Roost}, \texttt{ResCNN} and \texttt{RF\_Magpie} models.      


Table \ref{SA_table} summarises the MAEs of \texttt{Finder} and other structure-agnostic models on the same test set. It can be observed that \texttt{Finder} outperforms all other models irrespective of material property or dataset size. \texttt{Roost} produces impressive results too. We identify several focal points of our model that elevates its performance above that of \texttt{Roost}. First, our formula graph inherently encodes the stoichiometry whereas the representation of \texttt{Roost} requires propagating the fractional element weights to the message passing layer and the model performance might depend on where  and how the fractional weights are injected. Second, our edge predictive function fetches information from all other atoms in the formula graph to estimate edge attributes. This is inspired by the fact that the actual distance between two atoms in a crystal structure depends on other atoms in the crystal. Third, transformer-based self-attention component in our model includes trainable function approximators that enable abstracting the node attributes in different subspaces. This allows gaining a more vivid representation of the constituent atomic species before assigning an attention score to each of them. 

We further calculate the p-value between the MAE distributions of \texttt{Finder} and \texttt{Roost} to validate the statistical significance of our results (see Statistical Analysis section for more details). P-value is a measure of probability that an observed difference is merely due to random chance \cite{Ferreira2015}. Assuming that the cut-off value for the p-value to reject the null hypothesis is 0.05, we can conclude that the results of our model are statistically significant for all properties except for refractive index (see Table \ref{SA_table}).


\begin{table}[htb]
\centering

\begin{tabular}{ccccccc}
\hline
Property (unit)              & \texttt{Finder}     & \texttt{Roost}       & \texttt{ResCNN}     & \texttt{RF\_Magpie} & p-value$_{FR}$ & train-validation-test \\ \hline
$E_f$ (eV/atom)         & \textbf{0.0858(4)}  & 0.0913(8)   & 0.1131(11) & 0.1434(1)  & 0.0004         & 68699-14721-14722     \\
$E_{DFT}$ (eV/atom)      & \textbf{0.0896(1)}  & 0.0960(14)  & 0.1229(12) & 0.2058(1)  & 0.0017         & 68699-14721-14722     \\
$E_g$ (eV)          & \textbf{0.2911(9)}  & 0.3278(53)  & 0.3207(11) & 0.3321(3)  & 0.0003         & 68699-14721-14722     \\
n              & \textbf{0.1726(40)} & 0.1866(169) & 0.1975(21) & 0.3238(70) & 0.2345         & 3920-840-840          \\
$log(K_{VRH})$ (GPa) & \textbf{0.0835(6)}  & 0.0854(9)   & 0.0871(19) & 0.0934(2)  & 0.0372         & 7024-1506-1506        \\
$log(G_{VRH})$ (GPa) & \textbf{0.1153(14)} & 0.1226(19)  & 0.1314(16) & 0.1235(2)  & 0.0059         & 6699-1445-1441       \\ \hline
\end{tabular}
\caption{MAEs of structure-agnostic models in predicting six benchmark properties. The results show the mean and the standard deviation (in parenthesis) of MAE for three repeated runs with randomly initialised models. Same training, validation and test sets are used in evaluating all models. The number of samples in each set is shown in the last column. The best performing model is indicated in bold. p-value$_{FR}$ < 0.05 reflects that the difference between the results of \texttt{Finder} and \texttt{Roost} are statistically significant. The training progress comparison of \texttt{Finder} and \texttt{Roost} is shown in Supporting Figure S2.}
\label{SA_table}
\end{table}



Notably, \texttt{ResCNN} displays good MAE values despite operating on extremely simple material descriptors that only contain element fractions. This is possibly because standard convolutional layers are still remarkable feature extractors, and descriptor to property mapping function is potentially simplified by the use of a simple descriptor. Nevertheless, the performance of \texttt{ResCNN} and \texttt{RF\_Magpie} that use fixed-length descriptors is nearly always lower than that of graph-based models demonstrating the power of GNNs in representing diverse material compositions.

\begin{figure}[htb]
	\centering
	\includegraphics[width=0.8\textwidth]{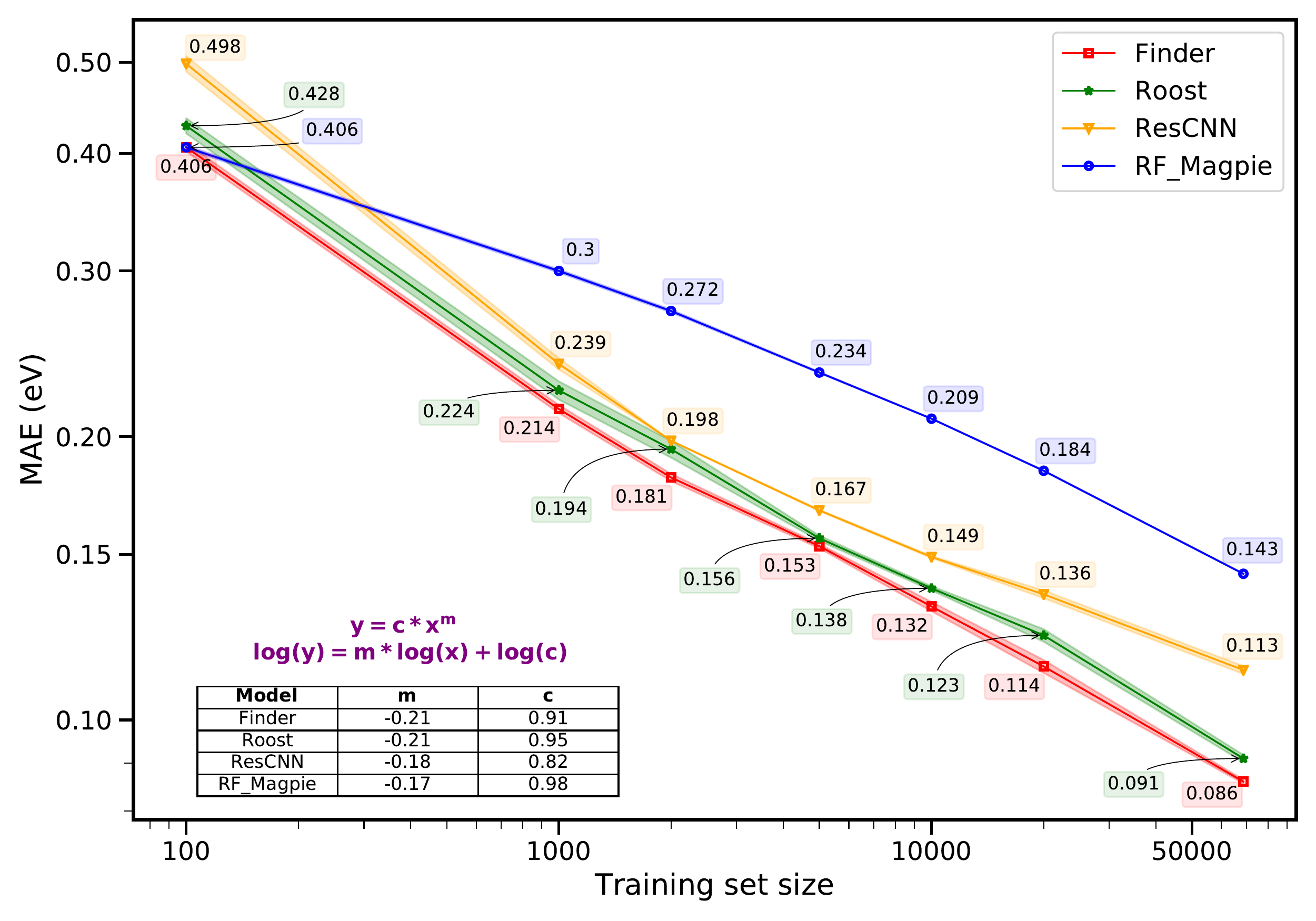}
\caption{Sample efficiency evaluation of structure-agnostic models on the MP formation energy dataset. Both axes are in log scale. The shaded region indicates the standard deviation of each model obtained from three repeated runs of randomly initialised models. The predictions in general abide by the power law \cite{Hestness}. By fitting to a power law function as shown in the figure, we obtain similar gradients for the error curves of \texttt{Finder} and \texttt{Roost} (\mbox{$m=-0.21$}). The absolute gradient is understandably smaller for \texttt{ResCNN} and \texttt{RF\_Magpie}.}
\label{figure3}
\end{figure}

Materials data, especially experimental measurements are often limited by size. This raises concerns on the competence of GNNs or ML models in general to learn from largely undersampled datasets and yet provide fairly accurate out-of-database predictions. To evaluate the sample efficiency of \texttt{Finder}, we observe its performance under different training set sizes. Figure \ref{figure3} depicts formation energy prediction MAE curves of all structure-agnostic models. \texttt{Finder} achieves the lowest error scores at all training set levels ranging from 10$^2$ to $\sim$ 7x10$^4$. Classical ML models that use explainable features such as \texttt{RF\_Magpie} are generally known to work well with small data. Despite inheriting from deep learning regime, our model starts outperforming \texttt{RF\_Magpie} as the training set size hits 10$^2$. While the MAE curve of \texttt{Finder} always hovers below that of \texttt{Roost}, the two error curves have a similar gradient.















\begin{figure}
	\centering
	\includegraphics[width=1\textwidth]{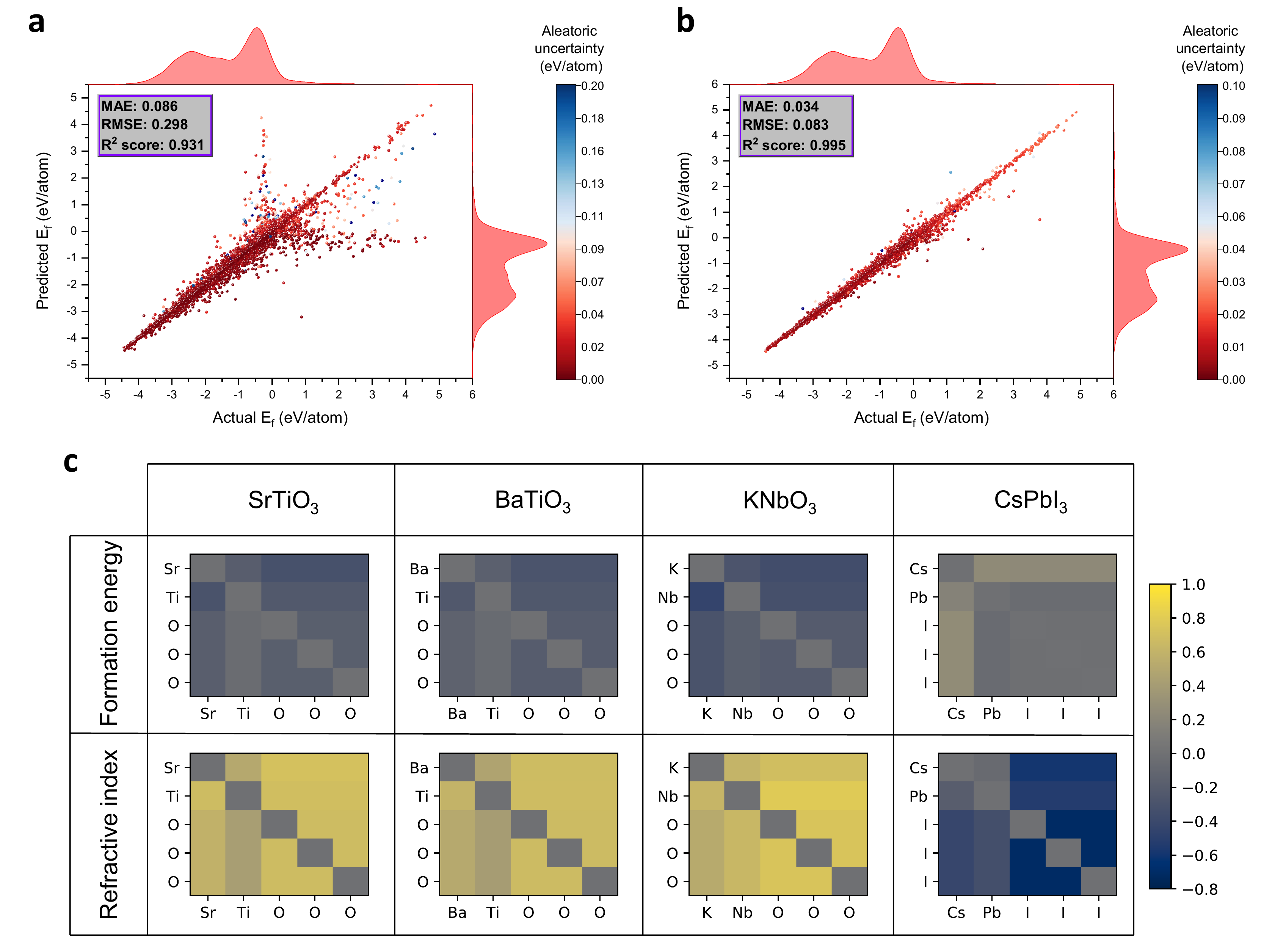}
\caption{Parity plots of structure-agnostic \texttt{Finder} (a) and structure-based variant (b) as obtained for the formation energy test set. The inclusion of spatial distances to the formula graph has significantly reduced both the error and the uncertainty in predictions. The marginal distributions of test data and the predictions are shown on the secondary axes. RMSE - root mean squared error, R$^2$ score - coefficient of determination. Parity plots for the other properties are included in Supporting Figure S3 and Figure S4. (c) Edge attribute matrices of perovskite materials \ch{SrTiO3}, \ch{BaTiO3}, \ch{KNbO3} and \ch{CsPbI3} visualised by probing $E_f$ and $n$ \texttt{Finder} models. The EAM is not necessarily symmetric because $e_{ij}^r$ is not always equal to $e_{ji}^r$ (see Equation \ref{edge_eqn}).}
\label{parity_plot}
\end{figure}

The $E_f$ parity plot shown in Figure \ref{parity_plot}a indicates that structure-agnostic \texttt{Finder} makes acceptable individual predictions, especially when the target value is negative. This is because computational materials  databases tend to report more stable materials that typically have a negative $E_f$. As expected, the aleatoric uncertainty of relatively inaccurate predictions is higher than that of the samples lying close to perfect prediction line.  


One key advancement of our network architecture is the simultaneous prediction of edge attributes from the associated node embeddings during training. We investigate whether these predicted edge values indeed capture some chemical or structural insights which are not explicitly fed to the model. Figure \ref{parity_plot}c presents edge attribute matrices (EAMs) of four well known perovskites, obtained from the final message passing layer of the trained $E_f$ and $n$ models, respectively. One might think of EAM as formula-domain relative of crystal-domain distance matrix. However, at it stands, such comparison is implausible because crystal structure may have several formula units. Nevertheless, from the $E_f$ model, we find that compositionally and structurally  similar materials such as \ch{SrTiO3} and \ch{BaTiO3} have similar EAMs. Intriguingly, compositionally different, yet structurally similar perovskite \ch{KNbO3} is found to have a comparable EAM to the ones above. The EAM of halide perovskite \ch{CsPbI3} is considerably different from its oxide counterparts. Consistent trend is observed from the refractive index model, yet resulting in different EAMs for the same material. Obviously, EAM entries are determined by the constituent element types and the data context. Although individual edge attributes carry no physical meaning, certain analogies between compositions can still be recovered from the EAM. Recently, we have shown that quantifying materials analogies can accelerate target driven discovery of materials \cite{Ihalage2021}. Such analogies rely on stoichiometry-derived global material embeddings. Incorporating  EAMs that reflect interactions between atoms adds another dimension for materials similarity analysis. 






\subsection{Structure-based model evaluation}

In our formula graph representation, shifting from structure-agnostic domain to structure-based domain is as simple as replacing the edge attributes with Gaussian-expanded atomic spacings and de-densifying the graph by connecting only the atoms arranged locally within a certain distance. This permits us to use the same message passing architecture, and any improvement in performance over structure-agnostic results is merely due to the addition of crystal structure, more specifically, the atomic spacings. In this section, we examine the structure-based variant of \texttt{Finder} with other materials graph networks such as \texttt{MEGNet} and \texttt{CGCNN}. Table \ref{SB_table} lists the MAEs of all structure-based models on the same benchmark datasets. \texttt{Finder} outperforms \texttt{CGCNN} in all properties and \texttt{MEGNet} in four out of six properties.


\begin{table}[htb]
\centering

\begin{tabular}{cccccc}
\hline
Property (unit)               & Finder     & MEGNet     & CGCNN      & p-value$_{FM}$ & train-validation-test \\ \hline
$E_f$ (eV/atom)          & \textbf{0.0342(3)}  & 0.0368(12) & 0.0425(6)  & 0.022          & 68699-14721-14722     \\
$E_{DFT}$ (eV/atom)      & 0.0351(1)  & \textbf{0.0332(12)} & 0.0890(17) & 0.0523         & 68699-14721-14722     \\
$E_g$ (eV)          & 0.2627(10) & \textbf{0.2609(7)}    & 0.2948(26) & 0.063          & 68699-14721-14722     \\
n              & \textbf{0.1554(33)} & 0.1654(39) & 0.2564(92) & 0.0270         & 3920-840-840          \\
$log(K_{VRH})$ (GPa) & \textbf{0.0728(3)}  & 0.0732(26) & 0.0829(19) & 0.8043         & 7024-1506-1506        \\
$log(G_{VRH})$ (GPa) & \textbf{0.1028(13}) & 0.1091(7)  & 0.1164(24) & 0.0018         & 6699-1445-1441       
\\ \hline
\end{tabular}
\caption{MAEs of structure-based models in predicting six benchmark properties. p-value$_{FM}$ < 0.05 means that the difference between the results of \texttt{Finder} and \texttt{MEGNet} are statistically significant.}
\label{SB_table}
\end{table}

Both \texttt{Finder} and \texttt{MEGNet} produce errors within  quantum chemical accuracy (1 kcal/mol or equivalently 43 meV/atom)  \cite{Bogojeski2020} for energy predictions. However, one should be mindful that the MP database contains many similar samples such as perovskites (that get split into both training and test set), and therefore the test errors reported here may not reflect the actual error of energy predictions for out-of-database samples including many hypothetical crystal structures. Figure \ref{parity_plot}b demonstrates the generalisability of our model for the $E_f$ test set. Notably, incorporating crystal structure lowers $E_f$ and $E_{DFT}$ prediction errors by over 60\%, a significant improvement from structure-agnostic results. $E_f$ and $E_{DFT}$ datasets are sufficiently sized and well assorted with 224 space group symmetries to facilitate a more granular learning of structural features. 


Surprisingly, structure-based band gap prediction only observes about 10\% reduction in error compared to its formula-only counterpart. This underpins the fact that band gap is a difficult quantity to predict even with modern DFT energy functionals \cite{Perdew96, Scuseriae2113648118}, demanding a certain degree of empiricism to counterpoise DFT errors \cite{Morales_Garc2017}. Noticing that overall, the error of structure-agnostic version of our model is similar to the error of \texttt{CGCNN} for all properties excluding formation energy is particularly encouraging.





\begin{figure}
	\centering
	\includegraphics[width=1\textwidth]{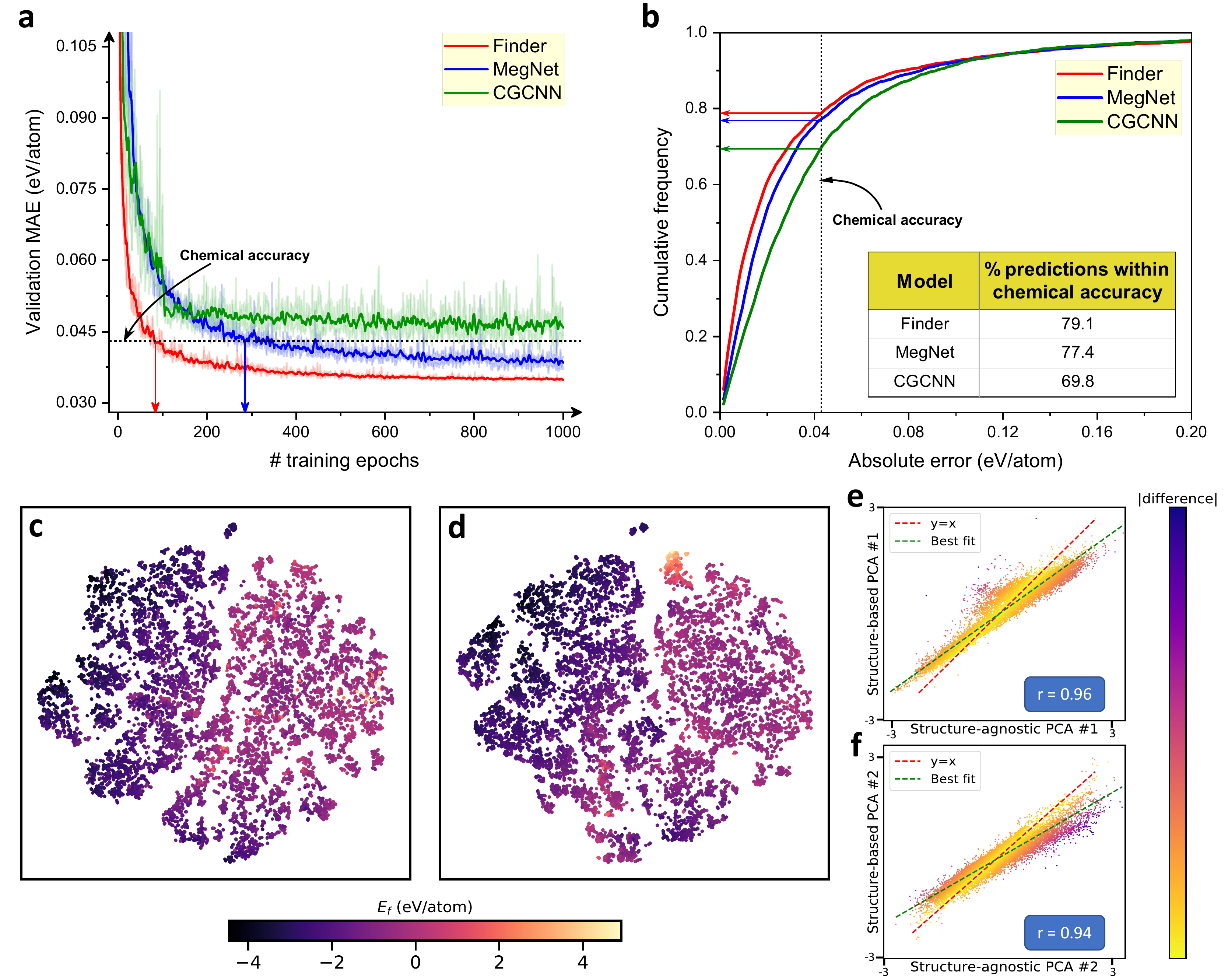}
\caption{Learning efficiency evaluation and t-SNE/PCA visualisations of material embeddings for the $E_f$ test set. (a) The training progress curves of \texttt{Finder}, \texttt{MEGNet} and \texttt{CGCNN}. Shaded region represents the standard deviation of MAEs. (b) Cumulative distribution plots illustrating the portion of predictions within a given absolute error. Scatter plots show t-SNE projections of the latent embeddings of crystals. The embeddings are taken from the final attention pooling layer of structure-agnostic (c) and structure-based (d) versions of our model. The scatters are color-coded according to the predicted $E_f$. Identical perplexity (30) and same random initialisations are used when creating the t-SNE plots. (e) Scatter plot shows the correlation between the first principal components of structure-agnostic and structure-based material embeddings, and (f) depicts the same trend between the second principle components. Each scatter in plots (c)-(e) represents a material from the $E_f$ test set.}
\label{le_cdf}
\end{figure}

The learning efficiency of a ML model is determined by how fast it reaches the convergence. An efficient learning model should ideally produce a lower error value relative to other competing models at any given point of time in the training process. Our model exhibits superior learning efficiency compared to \texttt{CGCNN} and \texttt{MEGNet} as corroborated by formation energy training curves in Figure \ref{le_cdf}a. \texttt{Finder} reaches chemical accuracy as quickly as in about 75 minutes (83 training epochs) whereas \texttt{MEGNet} takes about 500 minutes (285 epochs) to touch that level. \texttt{CGCNN} settles just around chemical accuracy after about 620 minutes (800 epochs). Figure \ref{le_cdf}b shows the cumulative distribution functions (CDFs) of three structure-based models. 79.1\% of predictions made by \texttt{Finder} is within the chemical accuracy, higher than \texttt{MEGNet} (77.4\%) and \texttt{CGCNN} (69.8\%). The CDF of our model stays above that of other models before all curves start to overlap near an error value of 0.12 eV/atom implying that all models find it equally difficult to predict the $E_f$ of the remaining portion of materials.


Figure  \ref{le_cdf}c and Figure  \ref{le_cdf}d refer to the t-distributed stochastic neighbour embedding (t-SNE) \cite{vandermaaten08a} visualisations of the internal representations of materials in the $E_f$ test set, assembled from structure-agnostic and structure-based \texttt{Finder} models, respectively. Interestingly, both latent maps resemble each other quite closely. Because t-SNE is a more visualisation oriented algorithm that involves non-linear projection, we  further perform principal component analysis (PCA) on the same data and observe a similar trend (PCA plots are provided in Supporting Figure S5). We investigate whether the location of the same material on both latent spaces is approximately similar by coupling the corresponding PCA components from the two domains. Figure \ref{le_cdf}e  displays the correlation between the first principal components of the material embeddings obtained with and without crystal structure. Figure \ref{le_cdf}f shows the same data for the second principal components. We find that structure-agnostic and structure-based  PCA components are highly correlated with a Pearson's correlation coefficient (r) of 0.96 between the first components and 0.94 between the second components. This means the two \texttt{Finder} models in general produce linearly related material embeddings where one is inferable from the other, although a few exceptions exist.

In structure-based domain, the latent vectors are expected to encode crystal structure details, and the materials in proximity are likely to be compositionally and structurally alike. This is quite compelling because it allows domain transferability at a reasonable fidelity. For example, structure-agnostic embedding of an undiscovered compound may be placed on the structure-based latent map and the neighbouring materials may be analysed to understand the possible crystal structure and the properties of the said hypothetical compound. This transferability was not possible between previous crystal graph and stoichiometry-only models due to contrasting ML model architectures and material descriptors. However, dedicated research is necessary to quantify the accuracy and validity of the specified strategy. Likewise, one has to perform such experiments in the original high-dimensional space rather than in the t-SNE or PCA reduced space for a more concrete analysis.







We note that training batch size has a considerable impact on the MAEs of all structure-based models. A batch size of 24 yields optimal results for both \texttt{Finder} and \texttt{CGCNN} in predicting formation energy, whereas \texttt{MEGNet} requires a relatively large batch size of 128 to stabilize training. Small batch sizes introduce noise in the error gradient estimation because model weights are updated more frequently. This might be desirable in some cases to circumvent local minima in the error surface. A batch size of 128 moderately increases formation energy MAEs of \texttt{Finder} and \texttt{CGCNN} to 0.0365 and 0.0455, respectively. Refractive index prediction uses an optimal batch size of 24 for all models and a default batch size of 128 operates well on all other properties.

It is worth mentioning that MAEs reported in this work for the structure-based models are slightly higher than those reported in original works \cite{Xie_Tian, Chen2019}. This is likely because we eliminate structural polymorphs from our databases. This intercepts the distinct advantage of seeing the same composition with slightly different properties triggered by the polymorphs, eventually increasing the prediction error by a small degree. Different sizes and random generation of training and test databases may also be a contributing factor. A standard materials database like the Matbench suite \cite{Dunn2020_automatminer} that also contains structural polymorphs may provide a wider platform to benchmark our approach against those from literature.





\subsection{Evaluation on Matbench v0.1 suite}

Matbench serves as a common test set for ML models and includes 13 materials property prediction tasks. Here, we select 8 regression tasks where the crystal structure is available. Two recent composition-only algorithms \texttt{AtomSets} \cite{Chen2021} and \texttt{CrabNet} \cite{Wang2021}, and two structure-based models \texttt{MEGNet} and \texttt{CGCNN} are selected for comparison. We follow the 5-fold nested cross validation (NCV) strategy with the same random seed variable recommended in the original study \cite{Dunn2020_automatminer} to evaluate all algorithms. Briefly, current NCV approach runs an outer test loop with 20\% test and 80\% training+validation data. For each outer NCV fold, there is an algorithm-dependent internal validation process. 

We note that non-graph models such as Automatminer \cite{Dunn2020_automatminer} and MODNet \cite{DeBreuck2021, De_Breuck_2021} that perform extensive hyperparameter tuning and feature selection steps at the inner loop of NCV typically outperform GNNs on small datasets. Once the internal optimisation is complete, such models are fit on the entire fold so that no validation data is left out. This approach is somewhat different from the common practice in GNNs where the validation data is indeed left out for model selection which actually reduces the amount of data available for training. This is the strategy used in ref. \cite{Dunn2020_automatminer} in evaluating \texttt{MEGNet} and \texttt{CGCNN} where the overall split for each fold is 60\% training, 20\% validation, and 20\% test. We use the same splitting ratio when evaluating our structure-based model to be consistent with previous results. Structure-agnostic \texttt{Finder} follows 72\% training, 8\% validation, and 20\% test portions criterion used in \texttt{AtomSets} \cite{Chen2021}.

Table \ref{matbench_table} shows the performance comparison on the Matbench suite. \texttt{Finder} achieves the best MAE scores in 5 out of 8 structure-agnostic tasks.  Especially, our model performs equally well on both small and large datasets. While \texttt{CrabNet} displays impressive results for small datasets (<$10^4$), our model outperforms \texttt{CrabNet} by up to 13\% as the database size reaches $\sim$ $10^5$.  \texttt{AtomSets} achieves the best MAE for the $perovskite$ formation energy dataset perhaps due to the use of a simple model and descriptor. Furthermore, \texttt{Finder} settles to better MAEs than alternative composition-only models such as \texttt{SkipAtom} \cite{antunes2021distributed} and \texttt{ElemNet} \cite{Jha2018}. \texttt{Finder} leads 5 out of 8 tasks in structure-based domain as well, consistent with our conclusions on the MP datasets with no polymorphs.

\begin{table}[htb]
\centering

\begin{tabular}{c|c|c|c|c|c|c|c}

\hline
\multirow{2}{*}{Property (unit)} & \multicolumn{3}{c|}{Structure-agnostic}                               & \multicolumn{3}{c|}{Structure-based}                               & \multirow{2}{*}{\begin{tabular}[c]{@{}c@{}}Dataset\\ size\end{tabular}} \\ \cline{2-7}
                                 & \multicolumn{1}{c|}{Finder} & \multicolumn{1}{c|}{AtomSets \cite{Chen2021}} & CrabNet \cite{Wang2021} & \multicolumn{1}{c|}{Finder} & \multicolumn{1}{c|}{MEGNet \cite{Chen2019}} & CGCNN \cite{Xie_Tian} &                                                                         \\ \hline
$jdft2d$ (meV/atom)                & 48                          & 52                            & \textbf{45.6}    & \textbf{46.1}                        & 55.9                        & 49.2   & 636                                                                     \\
$phonons$ (cm$^{−1}$)                   & \textbf{46.6}                        & 63                            & 55.1    & 50.7                        & \textbf{36.9}                        & 57.8   & 1265                                                                    \\
$dielectric$                       & \textbf{0.3204}                      & 0.36                          & \textbf{0.3234}  & \textbf{0.3197}                      & 0.478                       & 0.599  & 4764                                                                    \\
$log\_gvrh$ (GPa)                  & \textbf{0.0996}                      & 0.11                          & 0.1014  & 0.091                       & 0.0914                      & \textbf{0.0895} & 10987                                                                   \\
$log\_kvrh$ (GPa)                  & 0.0764                      & 0.08                          & \textbf{0.0758}  & \textbf{0.0693}                      & 0.0712                      & 0.0712 & 10987                                                                   \\
$perovskites$ (eV/atom)            & 0.645                       & \textbf{0.082}                         & 0.407   & \textbf{0.032}                       & 0.0417                      & 0.0452 & 18928                                                                   \\
$mp\_gap$ (eV)                     & \textbf{0.231}                       & 0.26                          & 0.266   & \textbf{0.219}                       & 0.235                       & 0.228  & 106113                                                                  \\
$mp\_e\_form$ (eV/atom)            & \textbf{0.0839}                      & 0.094                         & 0.0862  & 0.0342                      & \textbf{0.0327}                      & 0.0332 & 132752                                                                 
\end{tabular}
\caption{Performance comparison on the Matbench suite. The best performing models (within 1\% tolerance) in each domain are indicated in bold. The composition-only  results for \texttt{AtomSets} are taken from ref. \cite{Chen2021} and those for \texttt{MEGNet} and \texttt{CGCNN} are taken from ref. \cite{Dunn2020_automatminer}. \texttt{CrabNet}'s performance metrics are reported in Matbench leaderboard at \url{https://matbench.materialsproject.org/} (accessed on 1 March 2022). It should be noted that the datasets are used as-is (e.g. preprocessing such as removing duplicate compositions and outliers have not been applied)  for consistent comparison \cite{Dunn2020_automatminer}. $jdft2d$ - exfoliation energy \cite{Choudhary2020}; $phonons$ -  phonon DOS peak frequency \cite{Petretto2018}; $dielectric$ - refractive index \cite{Petousis2017}; $log\_gvrh$ -  log of shear moduli \cite{deJong2015}; $log\_kvrh$ -  log of bulk moduli \cite{deJong2015}; $perovskites$ - perovskite formation energy \cite{C2EE22341D}; $mp\_gap$ - band gap \cite{ONG2015209}; $mp\_e\_form$ - formation energy \cite{ONG2015209}.}
\label{matbench_table}
\end{table}

\subsection{Ablation experiments}

We perform ablation experiments to understand the causality of different components of our model and optimise its architecture. Table \ref{ablation_table} shows the results of ablation study on the $E_f$ database. While the number of message passing layers does not have a noticeable impact on the performance of structure-agnostic model, structure-based model observes a significant performance gain with multiple message passing layers compared to a solitary layer. Further increase of the number of message passing layers comes at the cost of degraded accuracy on small databases. As such, four message passing layers in composition-only domain increases the error in bulk modulus prediction by 5.3\% relative to the default architecture (shown in Supporting Table S1). 

Element embeddings of our model are transferred from ref. \cite{Tshitoyan2019}. Alternatively, we investigate one-hot element embeddings and observe a substantial dip in accuracy, particularly when the crystal structure is considered (see Model 1 in Table \ref{ablation_table}). This indicates node attributes that capture prior knowledge still help in navigating to a lower minimum in the error surface although this accuracy gap is expected to narrow down as the database size grows. Post-processing neural network is an essential component of our model as an exclusive message passing architecture inflates the error by up to 23\% (Model 2). We further note that including one standard convolutional layer in the post-processing network yields optimal results while adding more such layers impairs the performance. The number of rear dense layers and their widths are calibrated heuristically. 

In Model 3, we remove all residual connections and observe a substantial error increase in both structure-based and composition-only models. We then remove only the residual connections coming from message passing layers to post-processing network and keep the rest of the residual connections intact (Model 4). Somewhat unexpectedly, this reduces the error in structure-agnostic model while increasing the error in structure-based model. This means that while the reference architecture performs well overall, it is possible to achieve lower errors with domain-specific hyperparameter tuning. Note that both Model 3 and Model 4 reduce the number of $attn\_pool$ layers from $P$ to one, applied right after the last message passing layer. 

In order to verify that it is indeed the proposed formula graph representation that leads to improved performance, we replace the self-attention block in our message passing layers with a \texttt{Roost}-like soft-attention mechanism in Model 5. This makes material representation the most important distinction between Model 5 and  \texttt{Roost} as the rest of the functions such as local pooling and node update are mostly standard message passing operations. The MAE of Model 5 is 0.0871 whereas that of \texttt{Roost} is 0.0913, indicating that formula graph is a more complete representation of compositions. Slightly increased MAE in Model 5 compared to default \texttt{Finder} model implies that the proposed self-attention variant contributes to the performance. We observe this in Model 6 by removing the self-attention component from all message passing layers. Structure-agnostic MAE is significantly increased to 0.0931. We note that the self-attention part is vital in composition-only domain, while it has only a marginal effect in our structure-based results. Finally, we capture that it is the soft-attention component in global $attn\_pool$ layer that leads to improved performance compared to \texttt{MEGNet} and \texttt{CGCNN}. This can be observed in Model 7 as a global sum pooling layer without the attention section increases the structure-based error by over 12\%.










\begin{table}[htb]
\centering

\begin{tabular}{c|ccc|c|c|c|c|l|c|c|}
\cline{2-11}
                                         & \multicolumn{3}{c|}{\# Message passing layers}                     & \multirow{2}{*}{Model 1} & \multirow{2}{*}{Model 2} & \multirow{2}{*}{Model 3} & \multirow{2}{*}{Model 4} & \multirow{2}{*}{Model 5} & \multirow{2}{*}{Model 6} & \multirow{2}{*}{Model 7} \\ \cline{2-4}
                                         & \multicolumn{1}{c|}{1}      & \multicolumn{1}{c|}{2}      & 3      &                          &                          &                          &                          &                          &                          &                          \\ \hline
\multicolumn{1}{|c|}{Structure-agnostic} & \multicolumn{1}{c|}{0.0861} & \multicolumn{1}{c|}{\textbf{0.0858}} & 0.0867 & 0.0876                   & 0.1056                   & 0.0898                   & 0.0846                   & 0.0871                   & 0.0931                   & 0.0873                   \\ \hline
\multicolumn{1}{|c|}{Structure-based}    & \multicolumn{1}{c|}{0.0393} & \multicolumn{1}{c|}{\textbf{0.0342}} & 0.0342 & 0.0398                   & 0.0403                   & 0.0373                   & 0.036                    & 0.0366                   & 0.0345                   & 0.0386                   \\ \hline
\end{tabular}
\caption{$E_f$ MAEs of different model architectures considered in ablation study.  The reference model is indicated in bold. Other models differ from the default architecture of \texttt{Finder} as follows. Model 1 - uses one-hot node embeddings; Model 2 - post-processing network removed; Model 3 - all residual connections removed; Model 4 - only the residual connections coming from message passing layers removed; Model 5 - self-attention component in message passing layer replaced with a soft-attention mechanism. Model 6 - self-attention component removed from message passing layer; Model 7 - soft-attention component removed from global $attn\_pool$ layer.}
\label{ablation_table}
\end{table}

\subsection{Epsilon-near-zero materials discovery as an application of Finder}

Undoubtedly, the intimacy between chemical structures and graph neural networks produces excellent results in predicting various material properties. However, GNNs may still find it challenging to predict materials properties in the form of a complex function. This renders a multi-output regression problem. Recently, GNNs have been successfully applied to predict the absorption spectra of three-cation metal oxides \cite{Shufeng} and phonon density of states \cite{Chen_Zhantao}. In a similar vein, a multi-class classification GNN is implemented to predict protein functions \cite{Vladimir2021}. Here, we employ structure-agnostic \texttt{Finder} to predict frequency-dependent dielectric constant of inorganic compounds and eventually locate epsilon-near-zero (ENZ) candidates. ENZ materials possess a vanishingly small permittivity at a certain frequency that induces exceptional properties, some of which are still being experimented following theoretical predictions \cite{Marini2016}. While structural ENZ materials such as metamaterials have been extensively studied, they achieve ENZ behaviour only as an effective property occurring on wavelengths larger than the size of structural unit, not to mention the increased fabrication cost and complexity. Hence, there is a growing interest in natural materials that exhibit ENZ phenomena, especially with low dielectric loss. 

We extract a database of real ($\varepsilon_{re}$) and imaginary ($\varepsilon_{im}$) dielectric functions from the JARVIS repository. These are calculated using the OptB88vdW (OPT) DFT functional and shown to agree well with experimental data \cite{Choudhary2018}. Duplicate compositions have been removed from this database by selecting the most stable polymorph, resulting in a total of 12,353 materials. While the dielectric constant can be anisotropic, the quantities for different directions such as \textit{xx, yy, zz,} usually follow a similar trend and show equivalent resonance frequencies. Therefore, we only focus on the \textit{xx} direction. $\varepsilon_{re}$ and $\varepsilon_{im}$ databases are divided into 80\% training, 10\% validation and 10\% test sets, separately. We investigate the dielectric function prediction performance of \texttt{Finder} and \texttt{ResCNN} on the test set. Both models achieve respectable performance metrics with \texttt{Finder} outperforming \texttt{ResCNN} as expected (see Table \ref{difunc_table}). 

\begin{figure}
	\centering
	\includegraphics[width=1\textwidth]{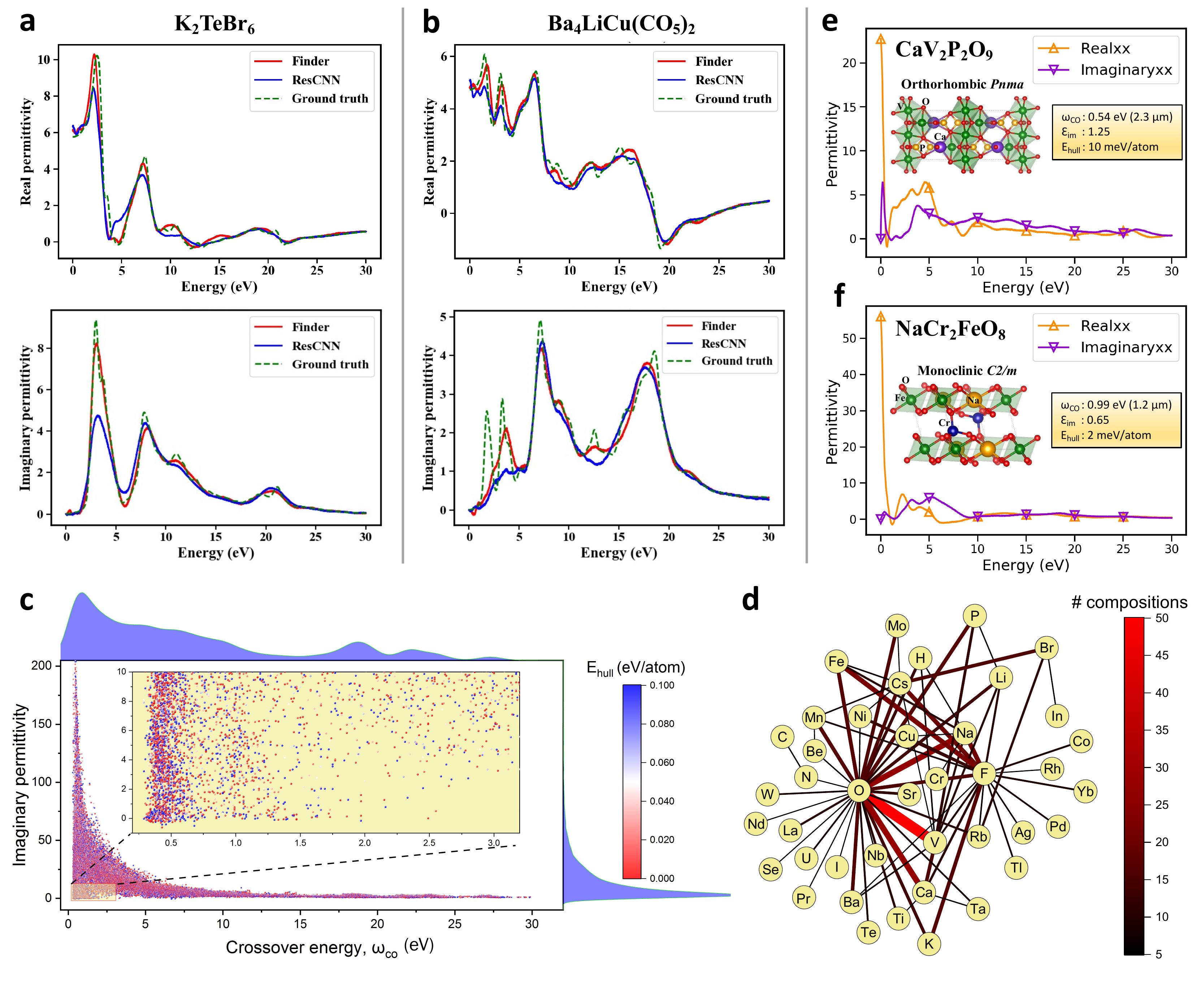}
\caption{Epsilon-near-zero materials discovery from the MP database. Frequency-dependent $\varepsilon_{re}$ and $\varepsilon_{im}$ functions of two representative materials \ch{K2TeBr6} (a) and \ch{Ba4LiCu(CO5)2} (b) as predicted by \texttt{Finder} and \texttt{ResCNN}. The x-axis represents the frequency in the units of photon energy. (c) Scatter plot shows the imaginary permittivity at the crossover energy point inferred from the predicted dielectric functions. (d) Network plot indicates element pairs that are present together in at least five predicted ENZ compositions. Each edge represents a pair of elements and the edge width is proportional to the number of element-pair appearances. (e) and (f) Show the predicted $\varepsilon_{re}$ and $\varepsilon_{im}$ of promising ENZ materials \ch{CaV2P2O9} and \ch{NaCr2FeO8} along with their crystal structure. 
}
\label{difunc_plot}
\end{figure}

Figure \ref{difunc_plot}a and Figure \ref{difunc_plot}b depict representative predictions for two selected materials from the test set. Note that \texttt{Finder} successfully captures several dielectric resonances of compositionally diverse materials (e.g. \ch{Ba2LiCu(CO5)2} as shown in Figure \ref{difunc_plot}b). What is more intriguing is that the predictions are based on a relatively small training database (9882 samples), and no structural information is incorporated in the process. Clearly, our model learns the complex mapping from composition to  dielectric function, bypassing the need for crystal structure and the computational complexity of DFT which is further intensified by accurate functionals such as OPT. 

The frequency range where $|\varepsilon_{re}| < 1$ is identified as the ENZ region. This is usually a narrow band positioned around the crossover energy point, the frequency at which the real part of permittivity crosses zero, $\omega_{co}$. While all metals achieve the ENZ condition at the bulk plasma resonance typically in the UV band, present studies are focussed on investigating ENZ materials in the NIR range that is close to the telecommunications wavelengths (1550 nm), as well as visible range that is directly accessible by optical experiments. Although the ENZ condition is only reliant on the real permittivity, large imaginary permittivity values, that is high loss, severely suppress the ENZ effect. In this work, we predict $\varepsilon_{re}$ and $\varepsilon_{im}$ functions of materials in the MP database and report low loss ENZ candidates ($\varepsilon_{im} < 2$) \cite{Kinsey2019} in NIR to UV band ($0.5 - 12.4$ eV) that are potentially stable (energy above convex hull, $E_{hull}<25$ meV). 


Figure \ref{difunc_plot}c depicts the inferred $\varepsilon_{im}$ vs $\omega_{co}$ dispersion. Evidently, the number of compositions with $\varepsilon_{im} < 2$ is extremely small compared to the composition space, making low loss ENZ materials discovery further challenging. We found 353 compositions from the MP database that satisfy above conditions. The candidate list covers 80 periodic table elements. Interestingly, the predicted ENZ compositions include alkaline earth metal vanadates such as \ch{Mg2V2O7} and \ch{Ca2V2O7} that relate to recently identified low loss ENZ materials, namely, CaVO3 and SrVO3 \cite{Zhang2016}. Strong electron-electron interactions present in such transition metal oxides can be capitalised to achieve ENZ condition in the visible spectrum. Unexpectedly, we found that 49 predicted ENZ materials contain vanadium and oxygen together, the most for any pair of periodic table elements. Other commonly appearing element pairs include \ch{Ca-O}, \ch{Na-O} and \ch{Fe-F} (see Figure \ref{difunc_plot}d). Linking these observations with already characterised ENZ correlated metals, we foresee vanadate compounds as an exciting class of materials for ENZ candidacy. Materials that feature a zero-permittivity wavelength in the NIR band are of great importance in telecommunications \cite{Kinsey2019}. We identify two potentially stable new compositions, \ch{CaV2P2O9} and \ch{NaCr2FeO8} that are predicted to exhibit low-loss ENZ properties in NIR region. Real and imaginary parts of the dielectric function as predicted by our model are shown in Figure \ref{difunc_plot}e and \ref{difunc_plot}f for the respective materials. $E_{hull}$ and $\varepsilon_{im}$ of both materials are well within the tolerable margins specified above. The full list of predicted ENZ compositions is provided as Supporting Data. We believe the search for functional ENZ materials has an expansive future especially considering the demonstrated potential of correlated metals.






\begin{table}[htb]
\centering
\begin{tabular}{c|cccc|cccc|}
\cline{2-9}
                             & \multicolumn{4}{c|}{$\varepsilon_{re}$}                                                                                & \multicolumn{4}{c|}{$\varepsilon_{im}$}                                                                           \\ \cline{2-9} 
                             & \multicolumn{1}{c|}{MAE}           & \multicolumn{1}{c|}{RMSE} & \multicolumn{1}{c|}{R$^2$ score} & MAD:MAE & \multicolumn{1}{c|}{MAE}           & \multicolumn{1}{c|}{RMSE} & \multicolumn{1}{c|}{R$^2$ score} & MAD:MAE \\ \hline
\multicolumn{1}{|c|}{Finder} & \multicolumn{1}{c|}{\textbf{0.69}} & \multicolumn{1}{c|}{3.05} & \multicolumn{1}{c|}{0.81}     & 4.1     & \multicolumn{1}{c|}{\textbf{0.72}} & \multicolumn{1}{c|}{3.18} & \multicolumn{1}{c|}{0.81}     & 5.2     \\ \hline
\multicolumn{1}{|c|}{ResCNN} & \multicolumn{1}{c|}{0.76}          & \multicolumn{1}{c|}{3.39} & \multicolumn{1}{c|}{0.82}     & 3.7     & \multicolumn{1}{c|}{0.77}          & \multicolumn{1}{c|}{3.45} & \multicolumn{1}{c|}{0.83}     & 4.8     \\ \hline
\end{tabular}
\caption{Performance metrics of \texttt{Finder} and \texttt{ResCNN} on frequency-dependent dielectric function prediction. Both models are trained with the $L_1$ loss. MAD stands for mean absolute deviation.}
\label{difunc_table}
\end{table}

\section{Discussion}

Both structure-based and structure-agnostic branches of materials property prediction come with inherent pros and cons. Current GNNs are limited to only one of the two branches owing to the little overlap between the material representations adopted in these two domains. In our unified approach, we aim not only to achieve state-of-the-art materials property prediction performance in both domains but also to enable transferability between composition and crystal structure embeddings that may lead to further research opportunities such as crystal structure prediction and structure prototype selection for DFT.


In support of this objective, we propose formula graph, a systematic generalisation of composition-only and crystal structure dependent material representations. Our intuition is to denote individual atoms in a chemical formula as nodes in a graph. The only decisive fragment between composition-only and structure-based formula graphs is the edge attribute which is readily available as atomic spacings for the latter and predicted during training for the former. We construct a self-attention driven message passing GNN and demonstrate that our model outperforms some previously reported models irrespective of the representation domain in predicting various material properties.

Moreover, our model displays better sample efficiency and learning efficiency compared to other models. This makes it a frontrunner for small data learning tasks that are abundant in materials science. Our deep learning baseline model produces respectable results in many tasks. We reckon standard CNNs still have some scope left in materials informatics, especially owing to the use of simple descriptors and easier implementation.

Finally, we expose \texttt{Finder} to a challenging task of predicting the frequency-dependent dielectric constant of inorganic compounds. Subsequently, we identify promising low loss ENZ materials that are of technological importance especially in optics and antenna engineering, demonstrating a real-world materials discovery application.

Our framework is not restricted to stoichiometric compounds. It can represent alloys, non-stoichiometric compounds or doped substances by converting the fractional element contributions to integer values. However, too small doping ratios multiply the size of formula graph and increase time and memory complexities, similar to how generating too large a supercell can make DFT calculations intractable. The same is true for alloys. As such, transforming a fractional formula of \ch{A_{0.33}B_{0.67}} to \ch{AB2} as its integer form makes more sense than a naïve conversion to \ch{A33B67}.

We view \texttt{Finder} as a potential distance matrix predictor that may help discover new crystal structures. This may be achieved, for example, by minimising the error between predicted edge attribute and the actual distance between corresponding atoms, instead of predicting a global material property. Alternatively, one might attempt to find a general mapping from EAM to distance matrix given $Z$. However, the existence of such a function is not a known priori. Yet another potential avenue of improvement is transfer learning from structure-based \texttt{Finder} to a structure-agnostic task as opposed to same domain transfer learning common in materials informatics. Nonetheless, these are recognised as future research directions. We believe domain invariant frameworks such as \texttt{Finder} that incorporate methodological successes from other disciplines including NLP and computer vision inaugurate a truly interdisciplinary avenue of research in materials science.

\section{Experimental Section}


We tune the architecture and hyperparameters of \texttt{Finder} to an adequate level by heuristically selecting a pool of hyperparameters that allows sufficient degrees of freedom yet remain computationally tolerable. We eventually settle to an architecture composed of two message passing layers followed by a post processing residual neural network with one convolutional-1D layer and four dense layers having 512, 1024, 1024 and 256 units, respectively. The element embeddings are adopted from ref. \cite{Tshitoyan2019} each having a dimension of 200. We realise that keeping the same dimension through our message passing layer improves the performance, although this is probed as a user-specified hyperparameter equivalent to the output shape of the preprocessing weight matrix $W_{int}$.

Function approximator networks $\phi_e$ and $\phi_m$ contain two hidden layers carrying 128 and 64 units. $F_Q$, $F_K$ and $F_V$ networks are all composed of individual weight matrices with a tunable output dimension defaulting to 200. Our global attention pooling component has pre- and post- processing layers each having 256-units. $L_2$ regularisation of 10$^{-6}$ is applied to all weights. We employ tensor clipping to evade exploding gradient problem \cite{pascanu13} which is one of the pitfalls of $L_1$ robust loss, despite its high empirical performance. 


Rectified linear units (ReLU) is used as hidden layer activation function that is changed to linear activation for the output layer. We use Adam optimiser with an initial learning rate of 3x10$^{-4}$ which is reduced by a factor of 0.999 at every iteration to allow finer convergence. Structure-agnostic models usually converge within 500 epochs with a batch size of 128 while structure-based models require about 1000 epochs to converge. Per-epoch timing for the $E_f$ database is about 35 seconds for the former and 55 seconds for the latter on an RTX 2080 Ti graphics processing unit. \texttt{Finder} is implemented in \textit{Keras} \cite{chollet2015keras} on top of \textit{Spektral} graph deep learning library \cite{Grattarola}.

We use a cutoff distance of  4 {\AA}  as used in \texttt{MEGNet} \cite{Chen2019} to derive the crystal graph. Structure-based edge attribute is of length 20 stemmed by the Gaussian expansion of spatial distance that takes the basis $exp(-(r-r_0)^2/\sigma^2)$ centred at 20 equidistant points between 0 to 5 where $\sigma = 0.5$ \cite{Chen2019}. \texttt{Roost}, \texttt{MEGNet} and \texttt{CGCNN} models are trained with recommended parameters from their respective repositories \cite{Goodall2020, Chen2019, Xie_Tian}. We investigate a small batch size of 24 in addition to the default value of 128 for all structure-based models. The models are trained for 1200 epochs or until stopped by an early stopping criterion. We note that because \texttt{MEGNet} discards crystal graphs with isolated atoms, its training set size in this work is slightly smaller (67814) relative to the full training set size (68699). However, we stick to the default cut-off distance of 4 {\AA} because increasing this value to 6 {\AA} downgrades the performance as found in the original work \cite{Chen2019} while multiplying the computational complexity by almost 3 times. According to the power law, such a marginal difference in training set sizes at that scale should have a negligible effect on the performance.


\texttt{RF\_Magpie} model adopts the implementation from \textit{scikit-learn}  \cite{pedregosa2011scikit} with default parameters and the \textit{Magpie} features are acquired from \textit{Matminer} package \cite{WARD201860}. \texttt{ResCNN} model is optimised to have four convolutional-1D layers containing 64, 128, 256 and 256 filters, respectively. A global max pooling layer is then placed to reduce the dimensionality and introduce local translation invariance. A set of postprocessing dense layers similar to that of \texttt{Finder} is appended with skip connections to complete our deep learning baseline architecture. The default parameters of \texttt{Finder} and \texttt{ResCNN} are given in Supporting Table S1 and Table S2, respectively. \\


\subsection{Statistical Analysis}

The training, validation and test sets of the MP datasets used in this work are kept intact for all algorithms to allow fair comparison. The target property values are z-score normalised based on training data as follows.

\begin{equation}
  \bar{x} = \frac{x - \mu_{tr}}{\sigma_{tr}}
\label{normalise}
\end{equation}

$\bar{x}$ is the normalised value of the original quantity $x$. $mu_{tr}$ and $\sigma_{tr}$ represent the mean and standard deviation of training data, respectively. The predictions are denormalised accordingly. The performance metrics MAE, RMSE, R$^2$ score and $r$ are obtained using  \texttt{scikit-learn} python package.

We perform a t-test to calculate the two-tailed p-value between two observed means $\bar{x}_1$ and  $\bar{x}_2$. The t-statistic ($t$) is calculated as follows. 


\begin{equation}
  SE(\bar{x}_1 - \bar{x}_2) = \sqrt{\left(\frac{(n_1-1)s_1^2+(n_2-1)s_2^2}{n_1+n_2-2}\right)\left(\frac{1}{n_1}+\frac{1}{n_2}\right)}
\label{se}
\end{equation}

\begin{equation}
  t = \frac{\bar{x_1}-\bar{x_2}}{SE(\bar{x}_1 - \bar{x}_2)}
\label{t_value}
\end{equation}

$n_1$ and $n_2$ are the number of samples in each group. In this experiment, $n_1 = n_2 = 3$. $s_1$ and $s_2$ are the two standard deviations. $SE(\bar{x}_1 - \bar{x}_2)$ denote the standard error. The significance level ($\alpha$) is set to 0.05. The p-value is calculated in python using \texttt{scipy.stats} package. The dielectric constant data is preprocessed to have a fixed dimension of 3000 points per sample. This corresponds to 3000 equally spaced points of photon energy ranging from 0 to 30 eV.

%

\medskip
\noindent \textbf{Supporting Information} \par 
\noindent
Supporting Information is available from the Wiley Online Library.

\medskip
\noindent \textbf{Acknowledgements} \par 

\noindent
The authors acknowledge funding received by The Institution of Engineering and Technology (IET) under the AF Harvey Research Prize. This work is supported in part by EPSRC Software Defined Materials for Dynamic Control of Electromagnetic Waves (ANIMATE) grant (EP/R035393/1).\\

\medskip

\noindent \textbf{Author Contributions} 

\noindent
A.I. designed the machine learning framework, performed data analysis and wrote the paper. Y.H. directed and coordinated the research. All authors discussed the results and reviewed the manuscript. \\

\medskip

\noindent \textbf{Competing Interests} 

\noindent
The authors declare that they have no competing interests.\\

\medskip

\noindent \textbf{Data availability} 

\noindent
The MP benchmark database curated in this study (\textit{MP\_2021\_July\_no\_polymorph}) is available at \url{https://doi.org/10.6084/m9.figshare.19308407}. Pre-trained \texttt{Finder} models are available at \url{https://doi.org/10.6084/m9.figshare.19308392.} The readers are directed to the original publications for Matbench \cite{Dunn2020_automatminer} and JARVIS dielectric constant data \cite{Choudhary2018, Choudhary2020}. The list of predicted ENZ materials are provided as Supporting Data. All other data needed to evaluate the findings of this study are present in the paper and/or the Supporting Information. \\

\medskip

\noindent \textbf{Code availability} 

\noindent
The codes required to reproduce the results of this study are available at \url{https://github.com/ihalage/Finder}. \\

%
\bibliographystyle{MSP}
\bibliography{article}

\begin{thebibliography}{10}
\providecommand{\url}[1]{\texttt{#1}}
\providecommand{\urlprefix}{URL }

\bibitem{Woodley2008}
S.~M. Woodley, R.~Catlow,
\newblock \emph{Nature Materials} \textbf{2008}, \emph{7}, 12 937.

\bibitem{HINE20091041}
N.~Hine, P.~Haynes, A.~Mostofi, C.-K. Skylaris, M.~Payne,
\newblock \emph{Computer Physics Communications} \textbf{2009}, \emph{180}, 7
  1041.

\bibitem{Bogojeski2020}
M.~Bogojeski, L.~Vogt-Maranto, M.~E. Tuckerman, K.-R. M{\"u}ller, K.~Burke,
\newblock \emph{Nature Communications} \textbf{2020}, \emph{11}, 1 5223.

\bibitem{Schmidt2019}
J.~Schmidt, M.~R.~G. Marques, S.~Botti, M.~A.~L. Marques,
\newblock \emph{npj Computational Materials} \textbf{2019}, \emph{5}, 1 83.

\bibitem{Reneaaq1566}
F.~Ren, L.~Ward, T.~Williams, K.~J. Laws, C.~Wolverton, J.~Hattrick-Simpers,
  A.~Mehta,
\newblock \emph{Science Advances} \textbf{2018}, \emph{4}, 4.

\bibitem{Weng2020}
B.~Weng, Z.~Song, R.~Zhu, Q.~Yan, Q.~Sun, C.~G. Grice, Y.~Yan, W.-J. Yin,
\newblock \emph{Nature Communications} \textbf{2020}, \emph{11}, 1 3513.

\bibitem{Raccuglia2016}
P.~Raccuglia, K.~C. Elbert, P.~D.~F. Adler, C.~Falk, M.~B. Wenny, A.~Mollo,
  M.~Zeller, S.~A. Friedler, J.~Schrier, A.~J. Norquist,
\newblock \emph{Nature} \textbf{2016}, \emph{533}, 7601 73.

\bibitem{Chen_Zhantao}
Z.~Chen, N.~Andrejevic, T.~Smidt, Z.~Ding, Q.~Xu, Y.-T. Chi, Q.~T. Nguyen,
  A.~Alatas, J.~Kong, M.~Li,
\newblock \emph{Advanced Science} \textbf{2021}, \emph{8}, 12 2004214.

\bibitem{Deringer}
V.~L. Deringer, M.~A. Caro, G.~Csányi,
\newblock \emph{Advanced Materials} \textbf{2019}, \emph{31}, 46 1902765.

\bibitem{Nagai2020}
R.~Nagai, R.~Akashi, O.~Sugino,
\newblock \emph{npj Computational Materials} \textbf{2020}, \emph{6}, 1 43.

\bibitem{Ward2016}
L.~Ward, A.~Agrawal, A.~Choudhary, C.~Wolverton,
\newblock \emph{npj Computational Materials} \textbf{2016}, \emph{2}, 1 16028.

\bibitem{Xie_Tian}
T.~Xie, J.~C. Grossman,
\newblock \emph{Phys. Rev. Lett.} \textbf{2018}, \emph{120} 145301.

\bibitem{Chen2019}
C.~Chen, W.~Ye, Y.~Zuo, C.~Zheng, S.~P. Ong,
\newblock \emph{Chemistry of Materials} \textbf{2019}, \emph{31}, 9 3564.

\bibitem{Ihalage2021}
A.~Ihalage, Y.~Hao,
\newblock \emph{npj Computational Materials} \textbf{2021}, \emph{7}, 1 75.

\bibitem{Yuan_Ruihao}
R.~Yuan, Y.~Tian, D.~Xue, D.~Xue, Y.~Zhou, X.~Ding, J.~Sun, T.~Lookman,
\newblock \emph{Advanced Science} \textbf{2019}, \emph{6}, 21 1901395.

\bibitem{Balachandran2018}
P.~V. Balachandran, B.~Kowalski, A.~Sehirlioglu, T.~Lookman,
\newblock \emph{Nature Communications} \textbf{2018}, \emph{9}, 1 1668.

\bibitem{Lookman2019}
T.~Lookman, P.~V. Balachandran, D.~Xue, R.~Yuan,
\newblock \emph{npj Computational Materials} \textbf{2019}, \emph{5}, 1 21.

\bibitem{Oviedo2019}
F.~Oviedo, Z.~Ren, S.~Sun, C.~Settens, Z.~Liu, N.~T.~P. Hartono, S.~Ramasamy,
  B.~L. DeCost, S.~I.~P. Tian, G.~Romano, A.~Gilad~Kusne, T.~Buonassisi,
\newblock \emph{npj Computational Materials} \textbf{2019}, \emph{5}, 1 60.

\bibitem{Ghosh_Kunal}
K.~Ghosh, A.~Stuke, M.~Todorović, P.~B. Jørgensen, M.~N. Schmidt, A.~Vehtari,
  P.~Rinke,
\newblock \emph{Advanced Science} \textbf{2019}, \emph{6}, 9 1801367.

\bibitem{Dinic}
F.~Dinic, K.~Singh, T.~Dong, M.~Rezazadeh, Z.~Wang, A.~Khosrozadeh, T.~Yuan,
  O.~Voznyy,
\newblock \emph{Advanced Functional Materials} \emph{n/a}, n/a 2104195.

\bibitem{Griffin}
L.~A. Griffin, I.~Gaponenko, N.~Bassiri-Gharb,
\newblock \emph{Advanced Materials} \textbf{2020}, \emph{32}, 38 2002425.

\bibitem{Ziletti2018}
A.~Ziletti, D.~Kumar, M.~Scheffler, L.~M. Ghiringhelli,
\newblock \emph{Nature Communications} \textbf{2018}, \emph{9}, 1 2775.

\bibitem{C9TC06073A}
N.~Liu, A.~Ihalage, H.~Zhang, H.~Giddens, H.~Yan, Y.~Hao,
\newblock \emph{J. Mater. Chem. C} \textbf{2020}, \emph{8} 10352.

\bibitem{Batra2021}
R.~Batra, L.~Song, R.~Ramprasad,
\newblock \emph{Nature Reviews Materials} \textbf{2021}, \emph{6}, 8 655.

\bibitem{Gilmer}
J.~Gilmer, S.~S. Schoenholz, P.~F. Riley, O.~Vinyals, G.~E. Dahl,
\newblock In \emph{Proceedings of the 34th International Conference on Machine
  Learning - Volume 70}, ICML'17. JMLR.org, \textbf{2017} 1263–1272.

\bibitem{Scutt}
K.~T. Schütt, H.~E. Sauceda, P.-J. Kindermans, A.~Tkatchenko, K.-R. Müller,
\newblock \emph{The Journal of Chemical Physics} \textbf{2018}, \emph{148}, 24
  241722.

\bibitem{Choudhary2021}
K.~Choudhary, B.~DeCost,
\newblock \emph{npj Computational Materials} \textbf{2021}, \emph{7}, 1 185.

\bibitem{omee2021scalable}
S.~S. Omee, S.-Y. Louis, N.~Fu, L.~Wei, S.~Dey, R.~Dong, Q.~Li, J.~Hu,
\newblock Scalable deeper graph neural networks for high-performance materials
  property prediction, \textbf{2021}.

\bibitem{Fung2021}
V.~Fung, J.~Zhang, E.~Juarez, B.~G. Sumpter,
\newblock \emph{npj Computational Materials} \textbf{2021}, \emph{7}, 1 84.

\bibitem{D0CP01474E}
S.-Y. Louis, Y.~Zhao, A.~Nasiri, X.~Wang, Y.~Song, F.~Liu, J.~Hu,
\newblock \emph{Phys. Chem. Chem. Phys.} \textbf{2020}, \emph{22} 18141.

\bibitem{Johannes}
J.~Klicpera, J.~Gro{\ss}, S.~G{\"{u}}nnemann,
\newblock \emph{CoRR} \textbf{2020}, \emph{abs/2003.03123}.

\bibitem{Park_iCGCNN}
C.~W. Park, C.~Wolverton,
\newblock \emph{Phys. Rev. Materials} \textbf{2020}, \emph{4} 063801.

\bibitem{Qiao}
Z.~Qiao, M.~Welborn, A.~Anandkumar, F.~R. Manby, T.~F. Miller,
\newblock \emph{The Journal of Chemical Physics} \textbf{2020}, \emph{153}, 12
  124111.

\bibitem{Bart}
A.~P. Bart\'ok, R.~Kondor, G.~Cs\'anyi,
\newblock \emph{Phys. Rev. B} \textbf{2013}, \emph{87} 184115.

\bibitem{Ward_crystal}
L.~Ward, R.~Liu, A.~Krishna, V.~I. Hegde, A.~Agrawal, A.~Choudhary,
  C.~Wolverton,
\newblock \emph{Phys. Rev. B} \textbf{2017}, \emph{96} 024104.

\bibitem{Artrith}
N.~Artrith, A.~Urban, G.~Ceder,
\newblock \emph{Phys. Rev. B} \textbf{2017}, \emph{96} 014112.

\bibitem{Goodall2020}
R.~E.~A. Goodall, A.~A. Lee,
\newblock \emph{Nature Communications} \textbf{2020}, \emph{11}, 1 6280.

\bibitem{NIPS2017_3f5ee243}
A.~Vaswani, N.~Shazeer, N.~Parmar, J.~Uszkoreit, L.~Jones, A.~N. Gomez, L.~u.
  Kaiser, I.~Polosukhin \textbf{2017}, \emph{30}.

\bibitem{Wang2021}
A.~Y.-T. Wang, S.~K. Kauwe, R.~J. Murdock, T.~D. Sparks,
\newblock \emph{npj Computational Materials} \textbf{2021}, \emph{7}, 1 77.

\bibitem{Schmidt_attn}
J.~Schmidt, L.~Pettersson, C.~Verdozzi, S.~Botti, M.~A.~L. Marques,
\newblock \emph{Science Advances} \textbf{2021}, \emph{7}, 49 eabi7948.

\bibitem{wang_Buwei}
B.~wang, Q.~Fan, Y.~Yue,
\newblock \emph{Journal of Physics: Condensed Matter} \textbf{2022}.

\bibitem{Shufeng}
S.~Kong, D.~Guevarra, C.~P. Gomes, J.~M. Gregoire,
\newblock \emph{Applied Physics Reviews} \textbf{2021}, \emph{8}, 2 021409.

\bibitem{Choudhary2020}
K.~Choudhary, K.~F. Garrity, A.~C.~E. Reid, B.~DeCost, A.~J. Biacchi, A.~R.
  Hight~Walker, Z.~Trautt, J.~Hattrick-Simpers, A.~G. Kusne, A.~Centrone,
  A.~Davydov, J.~Jiang, R.~Pachter, G.~Cheon, E.~Reed, A.~Agrawal, X.~Qian,
  V.~Sharma, H.~Zhuang, S.~V. Kalinin, B.~G. Sumpter, G.~Pilania, P.~Acar,
  S.~Mandal, K.~Haule, D.~Vanderbilt, K.~Rabe, F.~Tavazza,
\newblock \emph{npj Computational Materials} \textbf{2020}, \emph{6}, 1 173.

\bibitem{Reshef2019}
O.~Reshef, I.~De~Leon, M.~Z. Alam, R.~W. Boyd,
\newblock \emph{Nature Reviews Materials} \textbf{2019}, \emph{4}, 8 535.

\bibitem{Kinsey2019}
N.~Kinsey, C.~DeVault, A.~Boltasseva, V.~M. Shalaev,
\newblock \emph{Nature Reviews Materials} \textbf{2019}, \emph{4}, 12 742.

\bibitem{Park2015}
J.~Park, J.-H. Kang, X.~Liu, M.~L. Brongersma,
\newblock \emph{Scientific Reports} \textbf{2015}, \emph{5}, 1 15754.

\bibitem{Korobenko2021}
A.~Korobenko, S.~Saha, A.~T.~K. Godfrey, M.~Gertsvolf, A.~Y. Naumov, D.~M.
  Villeneuve, A.~Boltasseva, V.~M. Shalaev, P.~B. Corkum,
\newblock \emph{Nature Communications} \textbf{2021}, \emph{12}, 1 4981.

\bibitem{Haim2013}
S.~Haim, O.~Kevin, W.~Z. Jing, S.~Alessandro, Y.~Xiaobo, Z.~Xiang,
\newblock \emph{Science} \textbf{2013}, \emph{342}, 6163 1223.

\bibitem{Kinsey15}
N.~Kinsey, C.~DeVault, J.~Kim, M.~Ferrera, V.~M. Shalaev, A.~Boltasseva,
\newblock \emph{Optica} \textbf{2015}, \emph{2}, 7 616.

\bibitem{KafaieShirmanesh2018}
G.~Kafaie~Shirmanesh, R.~Sokhoyan, R.~A. Pala, H.~A. Atwater,
\newblock \emph{Nano Letters} \textbf{2018}, \emph{18}, 5 2957.

\bibitem{Andreas_residual}
A.~Veit, M.~J. Wilber, S.~J. Belongie,
\newblock \emph{CoRR} \textbf{2016}, \emph{abs/1605.06431}.

\bibitem{Qimai}
Q.~Li, Z.~Han, X.~Wu,
\newblock \emph{CoRR} \textbf{2018}, \emph{abs/1801.07606}.

\bibitem{Kaixiong}
K.~Zhou, X.~Huang, Y.~Li, D.~Zha, R.~Chen, X.~Hu,
\newblock \emph{CoRR} \textbf{2020}, \emph{abs/2006.06972}.

\bibitem{Dunn2020_automatminer}
A.~Dunn, Q.~Wang, A.~Ganose, D.~Dopp, A.~Jain,
\newblock \emph{npj Computational Materials} \textbf{2020}, \emph{6}, 1 138.

\bibitem{Ferreira2015}
J.~C. Ferreira, C.~M. Patino,
\newblock \emph{Jornal brasileiro de pneumologia : publicacao oficial da
  Sociedade Brasileira de Pneumologia e Tisilogia} \textbf{2015}, \emph{41}, 5
  485, 26578145[pmid].

\bibitem{Hestness}
J.~Hestness, S.~Narang, N.~Ardalani, G.~F. Diamos, H.~Jun, H.~Kianinejad,
  M.~M.~A. Patwary, Y.~Yang, Y.~Zhou,
\newblock \emph{CoRR} \textbf{2017}, \emph{abs/1712.00409}.

\bibitem{Perdew96}
J.~P. Perdew, K.~Burke, M.~Ernzerhof,
\newblock \emph{Phys. Rev. Lett.} \textbf{1996}, \emph{77} 3865.

\bibitem{Scuseriae2113648118}
G.~E. Scuseria,
\newblock \emph{Proceedings of the National Academy of Sciences} \textbf{2021},
  \emph{118}, 35.

\bibitem{Morales_Garc2017}
{\'A}.~Morales-Garc{\'i}a, R.~Valero, F.~Illas,
\newblock \emph{The Journal of Physical Chemistry C} \textbf{2017}, \emph{121},
  34 18862.

\bibitem{vandermaaten08a}
L.~van~der Maaten, G.~Hinton,
\newblock \emph{Journal of Machine Learning Research} \textbf{2008}, \emph{9},
  86 2579.

\bibitem{Chen2021}
C.~Chen, S.~P. Ong,
\newblock \emph{npj Computational Materials} \textbf{2021}, \emph{7}, 1 173.

\bibitem{DeBreuck2021}
P.-P. De~Breuck, G.~Hautier, G.-M. Rignanese,
\newblock \emph{npj Computational Materials} \textbf{2021}, \emph{7}, 1 83.

\bibitem{De_Breuck_2021}
P.-P.~D. Breuck, M.~L. Evans, G.-M. Rignanese,
\newblock \emph{Journal of Physics: Condensed Matter} \textbf{2021}, \emph{33},
  40 404002.

\bibitem{antunes2021distributed}
L.~M. Antunes, R.~Grau-Crespo, K.~T. Butler,
\newblock Distributed representations of atoms and materials for machine
  learning, \textbf{2021}.

\bibitem{Jha2018}
D.~Jha, L.~Ward, A.~Paul, W.-k. Liao, A.~Choudhary, C.~Wolverton, A.~Agrawal,
\newblock \emph{Scientific Reports} \textbf{2018}, \emph{8}, 1 17593.

\bibitem{Petretto2018}
G.~Petretto, S.~Dwaraknath, H.~P.C.~Miranda, D.~Winston, M.~Giantomassi, M.~J.
  van Setten, X.~Gonze, K.~A. Persson, G.~Hautier, G.-M. Rignanese,
\newblock \emph{Scientific Data} \textbf{2018}, \emph{5}, 1 180065.

\bibitem{Petousis2017}
I.~Petousis, D.~Mrdjenovich, E.~Ballouz, M.~Liu, D.~Winston, W.~Chen, T.~Graf,
  T.~D. Schladt, K.~A. Persson, F.~B. Prinz,
\newblock \emph{Scientific Data} \textbf{2017}, \emph{4}, 1 160134.

\bibitem{deJong2015}
M.~de~Jong, W.~Chen, T.~Angsten, A.~Jain, R.~Notestine, A.~Gamst, M.~Sluiter,
  C.~Krishna~Ande, S.~van~der Zwaag, J.~J. Plata, C.~Toher, S.~Curtarolo,
  G.~Ceder, K.~A. Persson, M.~Asta,
\newblock \emph{Scientific Data} \textbf{2015}, \emph{2}, 1 150009.

\bibitem{C2EE22341D}
I.~E. Castelli, D.~D. Landis, K.~S. Thygesen, S.~Dahl, I.~Chorkendorff, T.~F.
  Jaramillo, K.~W. Jacobsen,
\newblock \emph{Energy Environ. Sci.} \textbf{2012}, \emph{5} 9034.

\bibitem{ONG2015209}
S.~P. Ong, S.~Cholia, A.~Jain, M.~Brafman, D.~Gunter, G.~Ceder, K.~A. Persson,
\newblock \emph{Computational Materials Science} \textbf{2015}, \emph{97} 209.

\bibitem{Tshitoyan2019}
V.~Tshitoyan, J.~Dagdelen, L.~Weston, A.~Dunn, Z.~Rong, O.~Kononova, K.~A.
  Persson, G.~Ceder, A.~Jain,
\newblock \emph{Nature} \textbf{2019}, \emph{571}, 7763 95.

\bibitem{Vladimir2021}
V.~Gligorijevi{\'{c}}, P.~D. Renfrew, T.~Kosciolek, J.~K. Leman, D.~Berenberg,
  T.~Vatanen, C.~Chandler, B.~C. Taylor, I.~M. Fisk, H.~Vlamakis, R.~J. Xavier,
  R.~Knight, K.~Cho, R.~Bonneau,
\newblock \emph{Nature Communications} \textbf{2021}, \emph{12}, 1 3168.

\bibitem{Marini2016}
A.~Marini, F.~J. Garc{\'i}a~de Abajo,
\newblock \emph{Scientific Reports} \textbf{2016}, \emph{6}, 1 20088.

\bibitem{Choudhary2018}
K.~Choudhary, Q.~Zhang, A.~C. Reid, S.~Chowdhury, N.~Van~Nguyen, Z.~Trautt,
  M.~W. Newrock, F.~Y. Congo, F.~Tavazza,
\newblock \emph{Scientific Data} \textbf{2018}, \emph{5}, 1 180082.

\bibitem{Zhang2016}
L.~Zhang, Y.~Zhou, L.~Guo, W.~Zhao, A.~Barnes, H.-T. Zhang, C.~Eaton, Y.~Zheng,
  M.~Brahlek, H.~F. Haneef, N.~J. Podraza, M.~H.~W. Chan, V.~Gopalan, K.~M.
  Rabe, R.~Engel-Herbert,
\newblock \emph{Nature Materials} \textbf{2016}, \emph{15}, 2 204.

\bibitem{pascanu13}
R.~Pascanu, T.~Mikolov, Y.~Bengio,
\newblock In S.~Dasgupta, D.~McAllester, editors, \emph{Proceedings of the 30th
  International Conference on Machine Learning}, volume~28 of \emph{Proceedings
  of Machine Learning Research}. PMLR, Atlanta, Georgia, USA, \textbf{2013}
  1310--1318,
\newblock \urlprefix\url{https://proceedings.mlr.press/v28/pascanu13.html}.

\bibitem{chollet2015keras}
F.~Chollet, et~al.,
\newblock Keras, \textbf{2015},
\newblock \urlprefix\url{https://keras.io}.

\bibitem{Grattarola}
D.~Grattarola, C.~Alippi,
\newblock \emph{Comp. Intell. Mag.} \textbf{2021}, \emph{16}, 1 99–106.

\bibitem{pedregosa2011scikit}
F.~Pedregosa, G.~Varoquaux, A.~Gramfort, V.~Michel, B.~Thirion, O.~Grisel,
  M.~Blondel, P.~Prettenhofer, R.~Weiss, V.~Dubourg, et~al.,
\newblock \emph{Journal of machine learning research} \textbf{2011}, \emph{12},
  Oct 2825.

\bibitem{WARD201860}
L.~Ward, A.~Dunn, A.~Faghaninia, N.~E. Zimmermann, S.~Bajaj, Q.~Wang,
  J.~Montoya, J.~Chen, K.~Bystrom, M.~Dylla, K.~Chard, M.~Asta, K.~A. Persson,
  G.~J. Snyder, I.~Foster, A.~Jain,
\newblock \emph{Computational Materials Science} \textbf{2018}, \emph{152} 60.

\end{thebibliography}




\end{document}